\documentclass[sigconf,screen=true,natbib=true]{acmart}

\AtBeginDocument{%
  }

\usepackage{booktabs}
\usepackage{graphicx}
\usepackage{acronym}
\usepackage{xcolor}
\usepackage{bbm}
\usepackage{mleftright}
\usepackage{subcaption}
\usepackage{listings}
\usepackage[inline]{enumitem}
\usepackage{numprint}
\usepackage{dsfont}
\usepackage{minted}
\usepackage{multicol}
\usepackage{multirow}

\definecolor{codegreen}{rgb}{0,0.6,0}
\definecolor{codegray}{rgb}{0.5,0.5,0.5}
\definecolor{codepurple}{rgb}{0.58,0,0.82}
\definecolor{backcolour}{rgb}{0.95,0.95,0.92}

\acrodef{LLM}{large language model}
\acrodef{IR}{information retrieval}
\acrodef{RecSys}{recommender systems}
\acrodef{DCG}{discounted cumulative gain}
\acrodef{PRP}{pairwise relevance prompting}
\acrodef{LTR}{learning-to-rank}
\acrodef{nDCG}{normalized DCG}
\acrodef{PL}{Plackett-Luce}
\acrodef{RL}{reinforcement learning}

\copyrightyear{2026}
\acmYear{2026}
\setcopyright{cc}
\setcctype{by}
\acmConference[ICTIR '26]{Proceedings of the 2026 International ACM SIGIR Conference on Innovative Concepts and Theories in Information Retrieval (ICTIR)}{July 25, 2026}{Melbourne, VIC, Australia}
\acmBooktitle{Proceedings of the 2026 International ACM SIGIR Conference on Innovative Concepts and Theories in Information Retrieval (ICTIR) (ICTIR '26), July 25, 2026, Melbourne, VIC, Australia}
\acmDOI{10.1145/3805713.3820396}
\acmISBN{979-8-4007-2600-2/2026/07}

\allowdisplaybreaks

\begin{document}

\title{Exposure-Based Reinforcement Learning to Rank}

\settopmatter{authorsperrow=4,printfolios=true}

\author{Harrie Oosterhuis}
\affiliation{
    \institution{University of Amsterdam}
    \city{Amsterdam}
    \country{NL}
}
\email{h.oosterhuis@uva.nl}
\authornote{Work done while Harrie Oosterhuis was working at Google DeepMind.}

\author{Rolf Jagerman}
\affiliation{
    \institution{Google DeepMind}
    \city{New York}
    \country{US}
}
\email{jagerman@google.com}

\author{Zhen Qin}
\affiliation{
    \institution{Google DeepMind}
    \city{New York}
    \country{US}
}
\email{zhenqin@google.com}

\author{Xuanhui Wang}
\affiliation{
    \institution{Google DeepMind}
    \city{Mountain View}
    \country{US}
}
\email{xuanhui@google.com}

\begin{abstract}
\Ac{RL} methods for \ac{LTR} can optimize (almost) any ranking goal, e.g., from \emph{precision} or \emph{\acl{DCG}} to \emph{fairness-of-exposure} or \emph{ranking distillation}. However, standard \ac{RL} is ineffective and computationally costly due to the enormous action space in \ac{LTR} settings.
Existing methods reach computational efficiency through \emph{custom gradient} computation algorithms, but they are very complex to implement and often clash with auto-differentiation.
Consequently, existing \ac{RL} for \ac{LTR} is not attractive to many practitioners.

We reconsider \ac{RL} for \ac{LTR} while actively avoiding reliance on custom gradients.
Contrary to the existing approaches, we focus on variance reduction and GPU computation. In doing so, we discover that high sample-efficiency can be reached through \emph{baseline corrections} and \emph{partial marginalization}.
Furthermore, we propose an abstraction that places gradient estimation behind a document-exposure distribution, this enables seamless \emph{plug-and-play} integration with auto-differentiation.
Thereby, one only has to implement a loss as a differentiable function of exposure and \ac{RL} for \ac{LTR} can optimize it using auto-differentiation.

Our experimental results reveal that our new exposure-based \ac{RL} for \ac{LTR} approach converges considerably faster and at significantly higher ranking performance than existing custom gradients, with no additional costs in computation time when using GPUs.
In contrast, existing custom gradients result in severe stability issues when converging over many epochs, which never occur for our methods.
Thus, we considerably improve \ac{RL} for \ac{LTR} methodology by increasing its effectiveness, efficiency, and ease of application.
\end{abstract}



\maketitle
\acresetall

\section{Introduction}

The optimization of ranking systems is a core problem in the fields of \ac{IR}~\citep{liu2009learning} and \ac{RecSys}~\citep{karatzoglou2013learning}.
Its main challenge is that sorting-based metrics result in flat and non-continuous loss functions that provide no useful gradients.
Correspondingly, the sub-field of \ac{LTR} is concerned with surrogate loss functions that can provide usable substitute gradients~\citep{jagerman2022rax}, e.g., through bounding functions~\citep{wang2018lambdaloss, agarwal2019general}, approximations~\citep{qin2010general, wu2009smoothing, grover2019stochastic, pobrotyn2021neuralndcg} or ad-hoc heuristics~\citep{burges2006learning, pasumarthi2019tf}.

Whilst most \ac{LTR} methods are specifically designed for traditional relevance metrics (i.e., \ac{DCG} or precision@K)~\citep{jarvelin2002cumulated, jagerman2022rax, wang2018lambdaloss}, a branch of \ac{LTR} methods based on \ac{RL}~\citep{sutton1999reinforcement} are uniquely applicable to a much larger class of exposure-based ranking metrics~\citep{oosterhuis2021plrank, oosterhuis2022, singh2019policy}.
Here exposure is analogous to the expected attention that a document receives from users due to a ranking model~\citep{diaz2020evaluating}.
The origin of this frame is in fairness-of-exposure metrics that aim to fairly distribute attention over sets of documents~\citep{singh2018fairness, singh2019policy, biega2018equity, diaz2020evaluating}.
However, traditional relevance metrics and certain ranking-distillation losses can also be posed as exposure-based.
Thereby, \ac{RL} for \ac{LTR} is considerably more versatile than most \ac{LTR} methodologies~\citep{oosterhuis2021plrank}.

Despite its versatility, \ac{RL} for \ac{LTR} is not straightforward to apply; \ac{RL} struggles with large action spaces which conflicts with the enormous number of possible permutations that can be chosen as rankings in the \ac{LTR} setting~\citep{bruch2020stochastic, oosterhuis2021plrank, sutton1999reinforcement}.
Even with the extensive use of sampling, standard \ac{RL} results in high variance and can have high computational costs.
Previous work has alleviated this issue by introducing \emph{PL-Rank} -- a computationally-efficient algorithm to estimate the policy gradient~\citep{oosterhuis2021plrank, oosterhuis2022}.
In practice, PL-Rank requires the implementation of a complicated custom gradient function, which makes it prone to implementation errors.
Additionally, PL-Rank was made to minimize computation time on CPUs, and in doing so, appears to sacrifice numerical stability, resulting in serious convergency issues.
Manual implementation of custom gradients also does not integrate well into auto-differentiation frameworks, which have become the standard for machine learning~\citep{jagerman2022rax, paszke2019pytorch, bradbury2018jax, abadi2016tensorflow}.

In this work, we reconsider \ac{RL} for \ac{LTR} while focusing on GPU computation and avoiding the usage of custom gradients.
Instead, we consider \ac{RL} techniques such as \emph{baseline corrections}~\citep{Kool2019Buy4R, sutton1999reinforcement} and \emph{partial marginalization} to increase sample-efficiency.
Furthermore, we propose an estimator on the exposure distribution~\citep{diaz2020evaluating, singh2019policy} instead of the final metric/loss; in implementation, this acts as an abstraction that places the \ac{RL} policy gradient estimation behind the exposure distribution.
Thereby, it seamlessly integrates with auto-differentiation in a \emph{plug-and-play} manner for any exposure-based loss function, as practitioners simply only have to implement a loss as a differentiable function over the exposure distribution.
Thus, we avoid the complicated implementation of custom gradients, by abstracting complexity away behind exposure distributions.

Our experimental results reveal that, for traditional \ac{LTR} metric optimization, our exposure-based estimator has higher sample-efficiency and better convergence than previous methods and other estimators~\citep{oosterhuis2021plrank, oosterhuis2022, sutton1999reinforcement}.
The additional computational costs of our methods over existing custom gradients is inconsequential on GPUs and easily outweighed by their faster convergence. 
For non-conven\-tional losses, i.e., for fairness and distillation, the exposure-based estimator provides considerably easier implementation and without sacrificing effectiveness or efficiency.
Thus our contributions make \ac{RL} for \ac{LTR} more effective whilst staying efficient and also greatly increasing their ease of application to exposure-based metrics.

\section{Related Work}

Traditional \ac{LTR} methods optimize deterministic ranking models for ranking metrics and are classically categorized as pointwise, pairwise or listwise methods~\citep{liu2009learning, karatzoglou2013learning, jagerman2022rax}.
\emph{Pointwise} methods consider ranking scores as individual independent predictions~\citep{fuhr1989optimum, chu2005gaussian, crammer2001pranking};
\emph{pairwise} methods consider pairs of documents and aim to minimize incorrect predicted inversions in their order~\citep{joachims2002optimizing, bartell1995learning, burges2005ltr, burges2010ranknet};
and \emph{listwise} methods execute a complete ranking operation to compute their loss or gradient~\citep{burges2010ranknet, burges2006learning, cao2007listwise, xia2008listwise, wang2018lambdaloss}.
Recently, \citet{oosterhuis2022} proposed several changes to this classical categorization; they argue that categories should be allowed to overlap (e.g., methods can be both pairwise and listwise~\citep{burges2006learning, bruch2019metric, wang2018lambdaloss}) and made the case for several new categories: \emph{metric-based} methods that aim to optimize a specific ranking metric~\citep{bruch2019metric, burges2006learning, taylor2008softrank, wang2018lambdaloss}; \emph{sample-approximations} for methods that rely on probabilistic sampling to approximate ranking policies or gradients~\citep{bruch2020stochastic, oosterhuis2021plrank, oosterhuis2022, ustimenko2020stochasticrank} (e.g., policy-gradient methods in \ac{RL} for \ac{LTR}~\citep{bruch2020stochastic, oosterhuis2021plrank, oosterhuis2022, oosterhuis2018complex, diaz2020evaluating}); and \emph{exposure-based} methods that base their loss on exposure distributions over documents~\citep{oosterhuis2021plrank, oosterhuis2022, diaz2020evaluating}.

Numerous \ac{LTR} methods apply probabilistic ranking models to approximate deterministic counterparts during optimization, and they almost-unanimously apply Plackett-Luce (PL) ranking models~\citep{oosterhuis2021plrank, oosterhuis2022, diaz2020evaluating, cao2007listwise, xia2008listwise, luce2025individualchoice, plackett1975analysis}.
Additionally, many \ac{LTR} methods aim to optimize PL ranking models, e.g., \citet{bruch2020stochastic} argues that their stochasticity leads to more robust ranking performance.
Furthermore, their probabilistic behavior is useful for online and counterfactual evaluation~\citep{hofmann2011probabilistic, schuth2015multileave, oosterhuis2020evaluation}, exploration for click-based \ac{LTR}~\citep{oosterhuis2018ultr, oosterhuis-phd-thesis-2020, oosterhuis2023doubly} and fairness of exposure in rankings~\citep{oosterhuis2021plrank, singh2019policy, diaz2020evaluating}.
\citet{diaz2020evaluating} propose the concept of expected-exposure as the attention documents receive in expectation of a probabilistic ranking policy, and argue for ranking fairness based on expected exposure distributions; and also apply a PL ranking model in their experiments.
Finally, \citet{oosterhuis2021plrank, oosterhuis2022} introduced the \emph{PL-Rank} custom gradient algorithm to estimate the gradient of PL ranking models with the same computational complexity as sorting algorithms, for relevance and exposure-based fairness ranking metrics.

\section{Background}

\subsection{Traditional relevance ranking objectives}

Traditionally, \ac{LTR} aims to find ranking models that maximize the expected utility of rankings~\citep{liu2009learning}.
Let $y$ be a ranking of $D$ documents: $(y_1, y_2, \ldots, y_D)$, $q$ a query, $\theta_k \in \mathds{R}_{\geq0}$ position weights and  $\rho_{d \mid q} \in \mathds{R}_{\geq 0}$ document weights,
the utility of $y$ for $q$ is:
\begin{equation}
\textstyle
U(y \mid q) = \sum_{k=1}^K \theta_k \, \rho_{y_k \mid q}.
\end{equation}
Generally, $\rho_{d \mid q}$ captures the relevance of document $d$ to query $q$ by a numeric value, e.g., an expert label annotation or a conditional click-probability;
and $\theta_k$ captures the importance of the position $k$ by a numeric value (often $\theta_k \in [0,1]$), e.g., the ratio of users that examine a position~\citep{oosterhuis2023doubly, biega2018equity, joachims2017unbiased, diaz2020evaluating}, or matches prominent ranking metrics, i.e., \ac{DCG}: $\theta_k = (\log_2(k + 1))^{-1}$~\citep{jarvelin2002cumulated} or precision@K: $\theta_k = \mathds{1}[k < K]$.

We represent a ranking model as a stochastic policy $\pi$ with $\pi(y \mid q)$ being the probability that it provides ranking $y$ for query $q$.
The utility of a ranking model $\pi$ for a query $q$ is:
\begin{equation}
\textstyle
U(\pi \,|\, q)
= \mathbb{E}_{y\sim\pi(q)}\mleft[ U(y \,|\, q) \mright]
= \sum_{y \in \pi(q)}\! \pi(y \,|\, q) \sum_{k=1}^K \theta_k \rho_{y_k}
.
\label{eq:prelim:utility}
\end{equation}

Finally, the optimization goal is to maximize the utility of the ranking model over the true query distribution.
However, in practice, this is approximated with empirical risk minimization by averaging over a set of observed queries $Q$~\citep{vapnik1991principles}:
\begin{equation}
\textstyle
\mathcal{U} = \mathbb{E}_{q}\mleft[U(\pi \mid q)\mright]
\approx \frac{1}{|Q|} \sum_{q \in Q} U(\pi \mid q).
\end{equation}

\subsection{Exposure-based ranking objectives}
\label{sec:background:exposure}
Whilst traditional \ac{LTR} was mainly concerned with placing documents in order of relevance~\citep{liu2009learning, burges2006learning, wang2018lambdaloss}, a more recent perspective on \ac{LTR} regards it as optimizing a distribution of \emph{exposure} over documents~\citep{biega2018equity, singh2018fairness}.
This perspective is better aligned with stochastic ranking models and ranking fairness metrics~\citep{oosterhuis2021plrank, diaz2020evaluating}.
The exposure a document receives from ranking policy $\pi$ is the expected $\theta$ weight of the rank at which it is placed:
\begin{equation}
    \theta_{d \mid \pi,q}
    = \!\!\!\!\!\!
    \sum_{y\in \pi(q)} \!\!\!\! \pi(y \mid q) \sum_{k=1}^K \theta_k \mathds{1}\mleft[ y_k = d \mright]
    =
    \mathbb{E}_{y\sim\pi(q)}\mleft[ \theta_{\text{rank}(d \mid y)} \mright].
    \label{eq:prelim:exposure}
\end{equation}
Generally, the exposure $\theta_{d \mid \pi,q}$ should correspond to the amount of attention a document receives from users for query $q$~\citep{biega2018equity, singh2018fairness, diaz2020evaluating}.

The exposure-based perspective encapsulates traditional \ac{LTR}; let $\boldsymbol{\theta}$ and $\boldsymbol{\rho}$ be vectors of exposure and relevance, respectively:
\begin{equation}
    \begingroup
    \setlength\arraycolsep{2pt}
    \boldsymbol{\theta}_{\pi \mid q}^\top
    =
    \begin{bmatrix}
        \theta_{d_1 \mid \pi,q} &
        \dots &
        \theta_{d_D \mid \pi,q}
    \end{bmatrix}
    ,\quad
    \boldsymbol{\rho}_{q}^\top
    =
    \begin{bmatrix}
        \rho_{d_1 \mid q} &
        \dots &
        \rho_{d_D \mid q}
    \end{bmatrix}
    .
    \endgroup
\end{equation}
The utility of Equation~\ref{eq:prelim:utility} can then be formulated as a dot product:
\begin{equation}
\textstyle
U(\pi \mid q)
=
\boldsymbol{\rho}_{q}^\top \boldsymbol{\theta}_{\pi\mid q}
=
\sum_{i=1}^D  \theta_{d_i \mid \pi,q} \rho_{d_i \mid q}
.
\end{equation}
Moreover, exposure-based \ac{LTR} objectives can be characterized by a function $L$ that maps an exposure distribution for a query to a loss value.
The general loss is then the average value over a set of queries, for instance, for the traditional relevance objective:
\begin{equation}
\textstyle
\mathcal{L}(\pi) = \frac{1}{|Q|} \sum_{q \in Q} L(\boldsymbol{\theta}_{\pi \mid q}, \boldsymbol{\rho}_{q})
, \quad
    L_\text{rel}(\boldsymbol{\theta}, \boldsymbol{\rho}) = -\boldsymbol{\rho}^\top \boldsymbol{\theta}
    .
\end{equation}
Alternatively, \emph{exposure of fairness} metrics use functions commonly aim to reach an equal ratio between exposure and relevance, e.g., by penalizing the ratios between all pairs of documents~\citep{biega2018equity, singh2018fairness}:
\begin{equation}
    L_\text{frac-fair}(\boldsymbol{\theta}, \boldsymbol{\rho})
    =
    \frac{2}{D^2 - D}
    \sum_{i=1}^D \sum_{j=i+1}^D \mleft( \frac{ \theta_i }{\rho_i} - \frac{\theta_j}{\rho_j} \mright)^2.
    \label{eq:loss:fracfair}
\end{equation}
Because zero-relevance can result in undefined divisions, \citet{oosterhuis2021plrank} proposed a similar metric that avoids divisions:
\begin{equation}
    L_\text{prod-fair}(\boldsymbol{\theta}, \boldsymbol{\rho})
    =
    \frac{2}{D^2 - D}
    \sum_{i=1}^D \sum_{j=i+1}^D \mleft( \theta_i\rho_j  - \theta_j \rho_i \mright)^2.
    \label{eq:loss:prodfair}
\end{equation}
Let $\boldsymbol{\theta}'$ and $\boldsymbol{\rho}'$ indicate the vectors after normalization:
\begin{equation}
    \boldsymbol{\theta}' = \boldsymbol{\theta}/\lVert\boldsymbol{\theta}\lVert_1,
    \qquad
    \boldsymbol{\rho}' = \boldsymbol{\rho}/\lVert\boldsymbol{\rho}\lVert_1.
\end{equation}
This enables the application of many well-known divergence measures, for instance, a fairness metric based on KL-divergence~\citep{omer2021estimation}:
\begin{equation}
\textstyle
    L_\text{KL-fair}(\boldsymbol{\theta}, \boldsymbol{\rho}) = - \boldsymbol{\rho}'^\top \log( \boldsymbol{\theta}' \oslash \boldsymbol{\rho}' )
    =
    -\sum_{i=1}^D \rho_{i}' \log \mleft( \frac{ \theta_{i}' }{\rho_{i}'} \mright).
    \label{eq:loss:KLfair}
\end{equation}
Furthermore, exposure can be used to measure the difference between two ranking policies, which is at the core of ranking distillation~\citep{tang2018distillation, qin2023rdsuite, reddi2021rankdistil}.
Similarly, KL-divergence can be used for a distance loss to a desired exposure distribution $\boldsymbol{\theta^{*}}$:
\begin{equation}
\textstyle
    L_\text{KL-distill}(\boldsymbol{\theta}, \boldsymbol{\theta^{*}}) 
    \!=\!
    - \boldsymbol{\theta^{*}}'^\top \! \log( \boldsymbol{\theta}' \!\oslash \boldsymbol{\theta^{*}}')
   \!=\!
    -\!\sum_{i=1}^D \theta_i^{*\prime}\log\, \mleft( \frac{ \theta_i' }{\theta_i^{*\prime}} \mright).
    \label{eq:loss:KLdistill}
\end{equation}

\subsection{Plackett-Luce ranking models}

Similar to how the selection of actions by a \ac{RL} policy is generally modeled by the probabilities of a softmax function~\citep{sutton1999reinforcement}, probabilistic ranking policies are generally \ac{PL} ranking models which repeatedly sample from a softmax distribution without replacement~\citep{oosterhuis2021plrank, bruch2020stochastic, luce2025individualchoice, plackett1975analysis}.
The probability of placing a document in a \ac{PL} ranking model is the softmax probability over the logit scores $f(d \mid q)$ of the remaining available documents, i.e., for position $k$ and given partial ranking $y_{1:k-1}$:
\begin{equation}
\pi_f(y_k = d \,|\, y_{1:k-1})
=
\pi(d \,|\, y_{1:k-1})
=
\frac{
e^{f(d \mid q)}\mathds{1}\mleft[d \not\in y_{1:k-1}\mright]
}{\sum_{d \in D \setminus y_{1:k-1}}e^{f(d \mid q)}}
.
\end{equation}
Ranking is modeled by repeated sampling without replacement, therefore, the probability of a ranking is the product of each of its document placements:
\begin{equation}
\textstyle
\pi_f(y \mid q)
=
\pi(y)
=
\prod_{k=1}^K
\pi(y_k \mid y_{1:k-1}).
\end{equation}
Similarly, the probability of an incomplete ranking is:
\begin{equation}
\textstyle
\pi_f(y_{1:k} \mid  q)
=
\pi(y_{1:k})
=
\prod_{k'=1}^k
\pi(y_{k'} \mid y_{1:k'-1}).
\end{equation}
For brevity, we omit $f$ from our notation: $\pi_f = \pi$. %

An important property of Plackett-Luce ranking models is that rankings can be sampled efficiently by sorting the logits scores $f(d \mid q)$ after adding Gumbel noise to them~\citep{bruch2020stochastic}.
As a result, they work very well with sampling-based estimation techniques~\citep{oosterhuis2021plrank}.

\subsection{Policy-gradients and baseline corrections}
\label{sec:background:policygradient}
\emph{Contextual bandits} are a common \ac{RL} setting where policy can take an action $a$ given a context $x$ and receives an immediate reward $R(a,x)$ and contexts are independent of each other~\citep{sutton1999reinforcement, langford2007epoch}.
Thus, the expected reward of a policy is a nested expectation:
\begin{equation}
\textstyle
\mathcal{R}(\pi)
=
\mathbb{E}_{x}\big[\sum_{a \in \pi}\pi(a \mid x) R(a, x)\big]
=
\mathbb{E}_{x}[\mathbb{E}_{a \sim \pi}[R(a, x)]
.
\end{equation}
A common \ac{RL} method for optimizing such expected rewards is by taking the policy gradient~\citep{williams1992simple};
Let $f$ be the model underlying $\pi$:
\begin{equation}
\begin{split}
\frac{\delta \mathcal{R}(\pi)}{\delta f}
&=
\textstyle
\mathbb{E}_{x}\Big[\sum_{a \in \pi} \Big[\frac{\delta \pi(a \mid x)}{\delta f} \Big] R(a, x)\Big]
\\[-1ex]  &\approx
\textstyle
\frac{1}{N}\sum_{i=1}^N \Big[\frac{\delta \log \pi(a^{(i)} \mid x^{(i)})}{\delta f} \Big] R(a^{(i)}, x^{(i)}),
\end{split}
\end{equation}
using log-derivative trick:
$\frac{\delta \pi(a \mid x)}{\delta f} \!=\! \mathbb{E}_{a \sim \pi(x)}\!\big[\frac{\delta \log \pi(a \mid x)}{\delta f}\big]$,
for unbiased estimation from $N$ sampled contexts and actions.

The variance of policy-gradient estimators can pose a challenge, a popular variance reduction method is the baseline correction~\citep{greensmith2004variance}.
It stems from the fact that a (contextual) baseline function $b(x)$ can be subtracted from each individual reward whilst being added to the expected reward without introducing bias:
\begin{equation}
\textstyle
\mathcal{R}(\pi) = \mathbb{E}_{x}\big[\sum_{a \in \pi}\pi(a \mid x) \big(R(a, x) - b(x)\big)\big] + \mathbb{E}_{x}[b(x)].
\end{equation}
Since $b(x)$ is independent of $\pi$, the gradient can be estimated as:
\begin{equation}
\frac{\delta \mathcal{R}(\pi)}{\delta f}
 \approx
\frac{1}{N}\sum_{i=1}^N \Big[\frac{\delta \log \pi(a^{(i)} |\, x^{(i)})}{\delta f} \Big] \big(R(a^{(i)}\!\!, x^{(i)}) - b(x^{(i)})\big).
\end{equation}

The right choice of baseline $b$ can greatly reduce the variance of the gradient estimate, leading to faster and stabler convergence~\citep{gupta2024optimal, greensmith2004variance, dayan1991reinforcement}.
A common choice is the expected reward conditioned on $x$: $\mathbb{E}_{a \sim \pi(x)}[R(a, x)]$,
but this can be difficult if $x$ is often unique, thus another frequent choice is the unconditioned expected reward: $\mathbb{E}_{x}[\mathbb{E}_{a \sim \pi(x)}[R(a, x)]]$~\citep{willianms1988toward}.
With a leave-one-out mean~\citep{Kool2019Buy4R}:
\begin{equation}
\textstyle
    b(x^{(i)}) = \frac{1}{N-1} \sum^N_{j=1} \mathds{1}[i \not= j] R(a^{(j)}, x^{(j)}),
\end{equation}
an unbiased estimate of the difference $R(x,a) - b(x)$ is provided.

\subsection{PL-Rank: Fast but unstable custom gradient}

\citet{oosterhuis2021plrank, oosterhuis2022} proposed the PL-Rank algorithm for extremely fast computation of the policy-gradient of Plackett-Luce ranking models for ranking metrics.
Through a decomposition, PL-Rank-3~\citep{oosterhuis2022} computes a gradient from sampled rankings in linear time:
\begin{align}
PR_{y,k} &= \textstyle\sum_{k'=1}^{\min(k,K)} \theta_k \rho_{y_k}, 
&PR_{y,d} &= PR_{y,\text{rank}(d \mid y) + 1},
\nonumber \\[-0.6ex] 
RI_{y,k} &= \textstyle\sum_{k'=1}^{\min(k,K)} \frac{PR_{y,k'}}{\sum_{d\in D \setminus y_{1:k'-1}} e^{f(d \mid q)}},
&RI_{y,d} &= RI_{y,\text{rank}(d \mid y) + 1},
\nonumber \\[-0.6ex] 
DR_{y,k} &= \textstyle\sum_{k'=1}^{\min(k,K)}\frac{\theta_k}{\sum_{d\in D \setminus y_{1:k'-1}} e^{f(d \mid q)}},
&DR_{y,d} &= DR_{y,\text{rank}(d \mid y) + 1},
\nonumber \\[-0.6ex] 
&\hspace{-0.75cm}
\frac{\delta U(\pi \mid q) }{\delta f}
 \approx
 \frac{1}{N}
\textstyle \sum_{i=1}^N
 PR_{y^{(i)}\!\!,d} + e^{f(d\mid q)}\big(\rho_d DR_{y^{(i)}\!\!,d} - RI_{y^{(i)}\!\!,d}\big).
 \hspace{-5cm}&&
 \label{eq:PLRank}
\end{align}
The issue with computation following this decomposition is that the $RI$ and $DR$ values are not normalized, and can become enormously large or small, before being normalized in the final estimate.
This can lead to an extreme loss in precision in intermediate steps of the algorithm, resulting in errors in the final gradient estimate.
Also, PL-Rank does not integrate well with auto-differentiation software~\citep{jagerman2022rax, paszke2019pytorch, bradbury2018jax, abadi2016tensorflow} as it requires a custom gradient implementation.

\section{Method: RL for Traditional Relevance LTR}
\label{sec:method:relevance}

In this section, we build our \emph{marginalization} \ac{RL} approach for the traditional \ac{LTR} task by incrementally increasing complexity.
As a starting point, we note that exact computation of the utility $U(\pi \mid q)$ (Eq.~\ref{eq:prelim:utility}) involves a summation over all possible rankings, which is infeasible for non-trivial $K$ and $D$.
Fortunately, $U(\pi \mid q)$ is an expectation  that can be unbiasedly estimated through sampling, let $y^{(i)}$ be the $i$th ranking sampled from $\pi$ out of $N$ samples in total:
\begin{equation}
y^{(i)} \sim \pi(q), \qquad
U(\pi \mid q)
\approx
\textstyle
\frac{1}{N}\sum_{i=1}^N U(y^{(i)} \mid q).
\end{equation}
This is unbiased but $N$ should be large to account for variance.

\subsection{Standard policy-gradient approach}
If we directly apply the standard policy-gradient \ac{RL} approach as laid out in Section~\ref{sec:background:policygradient}~\citep{singh2018fairness, bruch2020stochastic}, then we obtain the following estimator:
\begin{equation}
\begin{split}
\frac{\delta }{\delta f}
U(\pi \mid q)
&\approx
\frac{1}{N}\sum_{i=1}^N \mleft(U(y^{(i)} \!\!\mid q) - b^{(i)}\mright) \mleft[ \frac{\delta }{\delta f} \log \pi(y^{(i)} \!\!\mid q) \mright],
\\[-1ex]
b^{(i)} &= \frac{1}{N-1}\sum_{j=1}^N U(y^{(j)} \mid q)\mathds{1}\mleft[i \not= j\mright].    
\end{split}
\label{eq:estimator:standard}
\end{equation}
We note that this estimator treats \emph{complete rankings} as actions.

\subsection{Treating rankings as placement trajectories}

\citet{oosterhuis2021plrank} notes that treating entire rankings as individual actions makes estimators unable to differentiate between utility added by different documents.
Consequently, if the first document is very relevant, then all subsequent documents are also rewarded for that.
Therefore, \citet{oosterhuis2021plrank} proposes to treat rankings as \emph{trajectories}: sequences of placement actions.
By reformulating utility:
\begin{equation}
U(\pi \,|\, q)
=
\sum_{k=1}^K \theta_k
\hspace{-0.9em} \sum_{y \in \pi(q)} \hspace{-0.7em}
\pi(y \,|\, q)
\rho_{y_k}
=
\sum_{k=1}^K \theta_k
\hspace{-0.9em}  \sum_{y_{1:k} \in \pi(q)} \hspace{-0.9em} 
\pi(y_{1:k} |\, q)
\rho_{y_k},
\end{equation}
we can derive the following matching estimator:
\begin{align}
\frac{\delta }{\delta f}
U(\pi \mid q)
&\approx
\frac{1}{N}\sum_{i=1}^N \sum_{k=1}^K \theta_k \mleft(\rho_{y_k^{(i)}} - b^{(i)}_k \mright)
\mleft[\frac{\delta }{\delta f}\log
\pi\mleft(y_{1:k}^{(i)} \mid q \mright)
\mright],
\nonumber \\[-0.7ex]
b^{(i)}_k &= \frac{1}{N-1}\sum_{j=1}^N \rho_{y_k^{(j)}} \, \mathds{1}\mleft[i \not= j\mright].
\label{eq:estimator:placement}
\end{align}
Here (baseline-corrected) rewards are only multiplied with the log-probability of the ranking up to their position ($\log \pi(y_{1:k} \mid q)$), thus the relevance of early documents does not increase the probability of later placements.
The main difference with the estimator of \citet{oosterhuis2021plrank} is that they do not utilize baseline corrections.

\subsection{Marginalizing over the first placement}
Since sampling is the reason for variance in our estimators, reducing its usage may reduce variance.
However, this is not straightforward as sampling is used because iterating over all rankings is infeasible.

We propose a middle-ground solution by avoiding sampling for computing the reward on the first position, while still relying on sampling for subsequent positions:
\begin{equation}
U(\pi \,|\, q)
\approx
\theta_1 \mleft( \sum_{d \in D}\!\! \pi(d \,|\, \emptyset) \rho_d \mright)
+
\frac{1}{N}\!\sum_{i=1}^N \sum_{k=2}^K\! \theta_k \rho_{\!\!y_k^{(i)}}.
\end{equation}
Thereby, we compute the exact expected reward on the first position with merely a summation over $D$ documents, while applying the placement approach for the remaining positions.
Thus, there is no variance stemming from the first position, as this value is no longer estimated.
For its gradient, we also do not have to rely on the log-derivative trick for first-position placements:
\begin{equation}
\begin{split}
&\frac{\delta }{\delta f} U(\pi \mid q)
\approx
\theta_1 \mleft( \sum_{d \in D} \mleft[\frac{\delta }{\delta f} \pi(d \mid \emptyset)\mright] (\rho_d - b_0^\prime)  \mright)
 \\[-0.5ex] & \qquad\qquad\;
+
\frac{1}{N} \sum_{i=1}^N \sum_{k=2}^K \theta_k \mleft(\rho_{\!\!y_k^{(i)}} - b_k^{ (i)}\mright)
\mleft[\frac{\delta }{\delta f}\log
\pi\mleft(y_{1:k}^{(i)} \mid q \mright)
\mright],
 \\[-1ex]
& b_0' = \sum_{d \in D} \pi(d \mid \emptyset) \rho_d
, \qquad
b^{(i)}_k =
\frac{1}{N-1}\sum_{j=1}^N \rho_{y_k^{(j)}} \, \mathds{1}\mleft[i \not= j\mright]
.
\end{split}
\label{eq:estimator:margfirst}
\end{equation}
Since this process is very similar to \emph{marginalizing} a distribution, we call it the \emph{marginalize-first} estimator.

\begin{figure}[t]
    \centering
    \includegraphics[width=\linewidth]{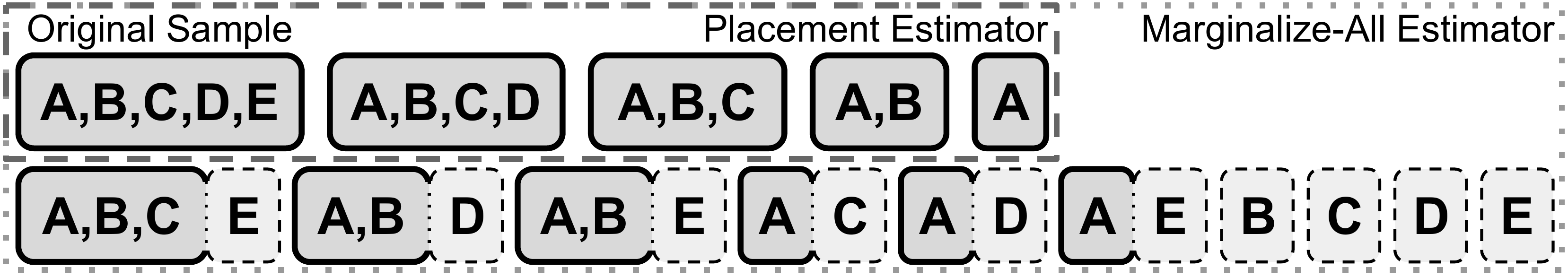}
    \vspace{-1.5\baselineskip}
    \caption{Example of a sampled ranking of a five-document collection; the partial rankings considered by the placement estimator; and the partial rankings that marginalize over the last placement considered by the marginalize-all estimator. }
    \label{fig:samplesoverview}
    \vspace{-1\baselineskip}
\end{figure}

\subsection{Marginalizing with all incomplete rankings}
\label{sec:method:allmargin}
Our next insight is that marginalization can be applied beyond the first position by conditioning on partial sampled rankings as prefixes.
For the second position, we can enumerate over all possible placements while conditioning on the first document of a sample:
\begin{equation}
\begin{split}
U(\pi \,|\, q)
&\approx
\theta_1 \mleft( \sum_{d \in D} \pi(d \,|\, \emptyset) \rho_d \mright)
 +
\theta_2 \mleft(\! \frac{1}{N} \! \sum_{i=1}^N \sum_{d \in D} \!\!\pi\mleft(d \,|\, y_{1:1}^{(i)}\mright) \rho_d  \mright)
 \\[-0.7ex]
&\quad +
\frac{1}{N}\!\sum_{i=1}^N \sum_{k=3}^K\! \theta_k \rho_{\!\!y_k^{(i)}}.
\end{split}
\end{equation}
By continuing this approach over all positions we obtain:
\begin{equation}
\begin{split}
U(\pi \mid q)
&\approx
\sum_{k=1}^K \theta_k \frac{1}{N} \sum_{i=1}^N \sum_{d \in D} \pi\mleft(d \mid y_{1:k-1}^{(i)} \mright) \rho_d 
\\[-0.7ex]
&=
\frac{1}{N} \sum_{i=1}^N \sum_{k=1}^K \theta_k \sum_{d \in D} \pi\mleft(d \mid y_{1:k-1}^{(i)} \mright) \rho_d,
\end{split}
\end{equation}
from which we get the following gradient estimator:
\begin{align}
&\frac{\delta }{\delta f}
U(\pi \mid q)
\approx
\frac{1}{N} \sum_{i=1}^N \sum_{k=1}^K\theta_k \sum_{d \in D}\Bigg(\mleft(\rho_d - b_k^{\prime(i)}\mright)
\nonumber\\[-1ex]&\qquad
\mleft(
\mleft[ \frac{\delta }{\delta f} \pi\mleft(d \,|\, y_{1:k-1}^{(i)} \mright)\mright] 
+ \pi\mleft(d \,|\, y_{1:k-1}^{(i)} \mright)\mleft[\frac{\delta }{\delta f} \log \pi\mleft(y_{1:k-1}^{(i)} \,|\, q \mright) \mright]
\mright)\Bigg),
\nonumber\\[-1ex]&\;
b^{\prime (i)}_k =
\frac{1}{N-1}\sum_{j=1}^N \mathds{1}\mleft[i \not= j\mright] \sum_{d \in D} \pi(d \mid y_{1:k-1}^{(j)}) \rho_d.
\label{eq:margallestimator}
\end{align}
Here the derivative is taken to the probability of single document placements and to the log probability of the sampled partial (prefix) rankings. %
We call this the \emph{marginalize-all} estimator.

By marginalizing over placements at all positions, the estimator considers more partial rankings than were sampled;
i.e., for each sampled ranking of length $K$, $D$ rankings of length one are considered, $D-1$ rankings of length two (as documents cannot be placed twice), $D-2$ rankings of length three, etc.
See Figure~\ref{fig:samplesoverview} for a visualization.
Therefore, for $N$ sampled (complete) rankings, the number of (non-unique) partial rankings considered is:
\begin{equation}
N(D + D-1 + D-2 + \ldots + D - (K-1)) = N\mleft(DK - \binom{K}{2}\mright).
\end{equation}
Through this strategy, more (partial) rankings are considered than those that are sampled, thereby, dependence on the sampling, and thus, variance from sampling is reduced.
This marginalization strategy involves extra computational costs, however, these are limited when computing on a GPU (Section~\ref{sec:method:implementation}).

This concludes our derivation of \ac{RL} for \ac{LTR} estimators for traditional relevance metrics, to the best of our knowledge, all these estimators are novel: the \emph{complete-actions} and \emph{placement} estimators merely due to their baseline-corrections, and the others also due to our novel \emph{marginalization} strategy.
Our proposed methodology reduces the variance of estimated gradients to improve optimization.

\section{Method: Exposure-Based RL for LTR}

This section adapts our novel marginalization method for traditional relevance \ac{LTR} (Section~\ref{sec:method:relevance}) to exposure-based \ac{LTR} (Section~\ref{sec:background:exposure}).

\subsection{Exposure-based estimation}
Instead of directly estimating the loss or utility of $\pi$, our exposure-based approach first estimates the exposure distribution of $\pi$.
In other words, we aim to estimate $\theta_{d|\pi,q}$ for each document $d$ (Eq.~\ref{eq:prelim:exposure}); we adapt the marginalization strategy over all positions from Section~\ref{sec:method:allmargin} to create the following estimator:
\begin{equation}
    \theta_{d|\pi,q} = \mathbb{E}_{y \sim \pi(q)}\big[ \theta_{\text{rank}(d \mid y)} \big]
    \approx 
    \frac{1}{N} \sum_{i=1}^N \sum_{k=1}^K \pi\mleft(d \mid y_{1:k-1}^{(i)} \mright) \theta_k.
    \label{eq:estimator:exposurevalue}
\end{equation}
Thereby, for each sampled ranking $y^{(i)}$ and each position $k$, we consider the probability that $d$ is placed after $y^{(i)}_{1:k-1}$ (i.e., the sampled documents up to $K$).
This leads to the new gradient estimator:
\begin{align}
\frac{\delta }{\delta f}
    \theta_{d|\pi,q}
    &\approx 
    \frac{1}{N} \sum_{i=1}^N \sum_{k=1}^K \mleft(\theta_k - b^{(i)}_d \mright)\bigg(\mleft[ \frac{\delta }{\delta f} \pi\mleft(d \mid y_{1:k-1}^{(i)} \mright)\mright]
    \nonumber \\[-1ex]
    & \qquad\qquad + 
    \pi\mleft(d \mid y_{1:k-1}^{(i)} \mright)\mleft[ \frac{\delta }{\delta f} \log \pi\mleft(y_{1:k-1}^{(i)} \mid q \mright) \mright]
    \bigg),
    \label{eq:estimator:exposure} \\[-1ex]
    b^{(i)}_d &=
\frac{1}{N-1}\sum_{j=1}^N \mathds{1}\mleft[i \not= j\mright] \sum_{k = 1}^K \pi\mleft(d \mid y_{1:k-1}^{(j)}\mright) \theta_k.
\nonumber
\end{align}
We call this the \emph{exposure-based} estimator.
Importantly, its baseline correction $b^{(i)}_d$ estimates the expected exposure of $d$; this contrasts with the corrections in Section~\ref{sec:method:relevance} as they estimate expected utility.

The exposure-based estimator only provides the gradient for the exposure distribution, in order to get the gradient of an exposure-based loss function one can apply the \emph{multivariable chain rule}~\citep{hughes2020calculus}:
\begin{equation}
\frac{\delta }{\delta f}
L(\boldsymbol{\theta}_{\pi \mid q})
= 
    \sum_{d \in D} 
    \bigg[\frac{\delta L(\boldsymbol{\theta}_{\pi \mid q})}{\delta \theta_{d \mid \pi,q}}
    \bigg]
\mleft[\frac{\delta \theta_{d \mid \pi,q}}{\delta f}
    \mright].
\label{eq:estimator:exposureloss}
\end{equation}
Thus, the gradient of the loss to the exposure distribution $\frac{\delta L(\boldsymbol{\theta}_{\pi | q})}{\delta \theta_{d|\pi,q}}$ -- which often can be computed exactly -- is separated from the gradient of the exposure distribution to the underlying ranking model $\frac{\delta \theta_{d|\pi,q}}{\delta f}$ -- which has to be estimated from samples.
Thereby, the estimator can be used for the optimization of any differentiable exposure-based ranking loss, objective or metric.

For example, if we chose our loss to match utility:
\begin{equation}
\frac{\delta }{\delta f}
U(\pi \,|\, q) \!
= \hspace{-0.5em}
    \sum_{d \in D} \hspace{-0.2em}
    \bigg[\frac{\delta U(\pi \,|\, q)}{\delta \theta_{d|\pi,q}}
    \bigg]
\mleft[\frac{\delta \theta_{d|\pi,q}}{\delta f}
    \mright] \!
= \hspace{-0.5em}
\sum_{d \in D} \! \rho_d 
\mleft[\frac{\delta }{\delta f}
    \theta_{d|\pi,q}
    \mright].
    \label{eq:exampleutilitygradient}
\end{equation}
We note that this estimator is very similar but not equivalent to the \emph{marginalize-all} estimator (Eq.~\ref{eq:margallestimator}), due to the difference in baseline corrections.
If we would disable these corrections then the estimators would be equivalent (i.e., if one sets all $b = 0$ then Eq.~\ref{eq:exampleutilitygradient} becomes equal to Eq.~\ref{eq:margallestimator}).
This difference means that both estimators are unbiased but may have differing variance.

\begin{listing}[t]
\begin{minted}[fontsize=\footnotesize]{python}
import jax.numpy as jnp
from jax.lax import stop_gradient
from utils import sample_rankings, cumlogsumexp
# input: [D] vector of scores, [K] vector of exposure, RNG key and N
# output: [D] vector of (estimated) expected exposure per document
def exposure(scores, rank_exposure, rng_key, n_samples):
  K = rank_exposure.shape[0]
  # creates [N, D] matrix of sampled rankings from PL ranking model
  rankings = sample_rankings(rng_key, n_samples, scores)
  ranked_scores = scores[rankings] 
  # [N, K] log denominator of placement prob. per position per sample
  log_denom = cumlogsumexp(ranked_scores, axis=1, reverse=True)[:,:K]
  # [N, K-1] log prob. of sampled ranking up to pos. k (not including)
  log_prefix = ranked_scores[:,:K-1] - log_denom[:,:K-1]
  log_prefix = jnp.cumsum(jnp.pad(log_prefix, ((0,0),(1,0))), axis=1)
  # [N, K, D] placement prob. of every document per position & sample
  prob_all = jnp.exp(scores[None,None,:] - log_denom[:,:,None])
  mask = jnp.zeros_like(prob_all, dtype=bool)
  mask = mask.at[jnp.arange(n_samples)[:,None], jnp.arange(1,K)[None,:],
                 rankings[:,:K-1]].set(True)
  mask = jnp.cumsum(mask, axis=1).astype(bool)
  prob_all = jnp.where(mask, 0, prob_all)
  # [N, K, D] loss value used for computing policy gradient
  loss_K_D = prob_all + stop_gradient(prob_all)*log_prefix[:,:,None]
  exposure_prob_prod = prob_all * rank_exposure[None,:,None]
  # [N, D] exposure of document per sample
  sample_exposure = jnp.sum(exposure_prob_prod, axis=1)
  # [D] expected exposure of document estimated from samples
  mean_exposure = jnp.mean(sample_exposure, axis=0)
  # [N, D] leave-one-out baseline corrections
  baseline = mean_exposure[None,:] - sample_exposure/n_samples
  baseline *= n_samples / (n_samples-1)
  # [N, K, D] baseline-corrected rewards for every doc., pos., sample
  rewards = stop_gradient(rank_exposure[None,:,None]-baseline[:,None,:])
  loss = jnp.mean(jnp.sum(rewards * loss_K_D, axis=1), axis=0)
  # trick: value is expected exposure but gradient comes from loss
  return stop_gradient(mean_exposure) + loss - stop_gradient(loss)
\end{minted}
\caption{Implementation of the exposure-based estimator.}
\label{listing:estimator}
\end{listing}

\begin{listing}[t]
\begin{minted}[fontsize=\footnotesize]{python}
exposures = exposure(scores, rank_exposure, rng_key, n_samples)
# relevances is a [D] vector of relevance labels per doc.
relevance_loss = -jnp.sum(exposures * relevances)

ratios = exposures / relevances
frac_fair_loss = jnp.mean((ratios[:,None] - ratios[None,:])**2)
frac_fair_loss *= ratios.size**2 / (ratios.size**2 - ratios.size)

norm_exp = exposures / jnp.sum(exposures)
norm_rel = relevances / jnp.sum(relevances)
KL_fair_loss = -jnp.sum(norm_rel * jnp.log(norm_exp / norm_rel))
\end{minted}
\caption{Implementation of multiple loss functions.}
\label{listing:losses}
\end{listing}

\subsection{Implementation with auto-differentiation}
\label{sec:method:implementation}

The exposure-based estimator can be implemented to integrate well into auto-differentiation software and make effective usage of GPU computation~\citep{jagerman2022rax}.
Listing~\ref{listing:estimator} displays a potential implementation in Python and JAX~\citep{bradbury2018jax}, where the \emph{exposure} function returns a vector of the (estimated) exposure distribution which is auto-differentiable by relying on the exposure-based policy-gradient estimator.
In other words, the implementation is integrated in such a way that the auto-differentiation of JAX automatically uses the exposure-based estimator for its gradient.
As a result, the gradient estimation is abstracted away and does not need to be considered for the subsequent implementation and optimization of exposure-based loss functions.
To illustrate this, Listing~\ref{listing:losses} provides example implementations of the traditional \ac{LTR} utility $U$ (Eq.~\ref{eq:prelim:utility}), $L_\text{frac-fair}$ (Eq.~\ref{eq:loss:fracfair}) and $L_\text{KL-fair}$ (Eq.~\ref{eq:loss:KLfair}).
By using the exposure distribution function, the implementation of each loss is no more than three lines.
Importantly, each of these losses is differentiable and ready to be used in stochastic gradient descent.
For comparison, \citet{oosterhuis2021plrank} proposes the usage of PL-Rank for fairness losses by first computing the gradient of the loss w.r.t.\ the exposure distribution and then using these values as the relevance labels for a PL-Rank gradient.
Therefore, our exposure-based estimator provides an enormous increase in \emph{ease-of-implementation} compared to the previous state-of-the-art.

Another advantage of our implementation is that all probabilities are computed in log-space with usage of \emph{logsumexp} operations~\citep{pierre2020accurately}.
Consequently, its computation is much more numerically stable that that of PL-Rank which uses un-normalized probabilities in several of its intermediate steps.
Nevertheless, the exposure based estimator has a computational complexity of $\mathcal{O}(N\times K\times D)$ which is worse than that of PL-Rank-3: $\mathcal{O}(N\times (K + K \log D))$~\citep{oosterhuis2022}.
Thus, it appears that a tradeoff is made between numerical stability and computational efficiency.
However, this tradeoff may not be so straightforward, since the exposure-based estimator consists only of matrix operations which work very well with GPU computation, whereas PL-Rank was not designed for GPU computations.
Thus, in practice, the decrease in computational efficiency may not be as impactful as one would expect \emph{prima facie}.

{
\renewcommand{\arraystretch}{0.3}
\setlength\tabcolsep{0.5pt}
\begin{figure*}[t]
    \centering
    \begin{tabular}{r l l l l}
    & \multicolumn{1}{c}{\hspace{2.5mm}\footnotesize N = 2}
    & \multicolumn{1}{c}{\!\footnotesize N = 10}
    & \multicolumn{1}{c}{\footnotesize N = 100\hspace{3mm}}
    & \multirow{3}{*}[-37.5pt]{\includegraphics[scale=0.4]{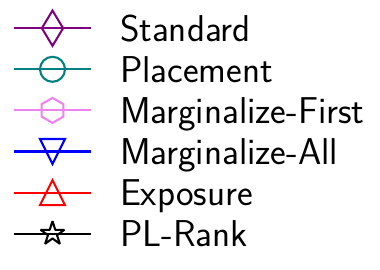}}
    \\
    \raisebox{2.5\normalbaselineskip}[0pt][0pt]{\rotatebox[origin=c]{90}{
\footnotesize MSLR - NDCG@10
}}
        &
        \includegraphics[scale=0.29]{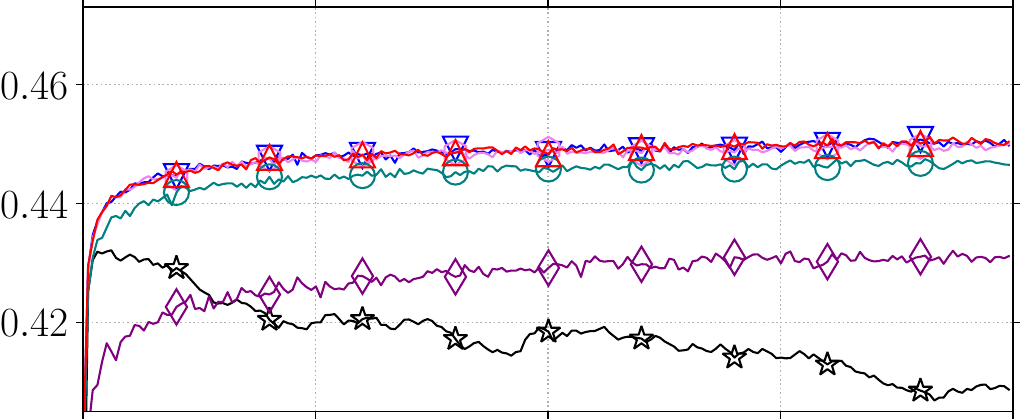}%
        \hspace{0.15mm}%
        &
        \hspace{0.1mm}%
        \includegraphics[scale=0.29]{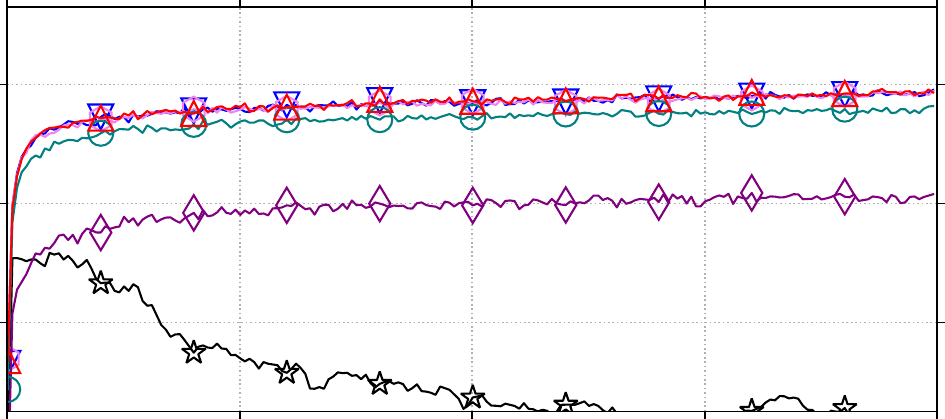}
        &
        \hspace{0.1mm}%
        \includegraphics[scale=0.29]{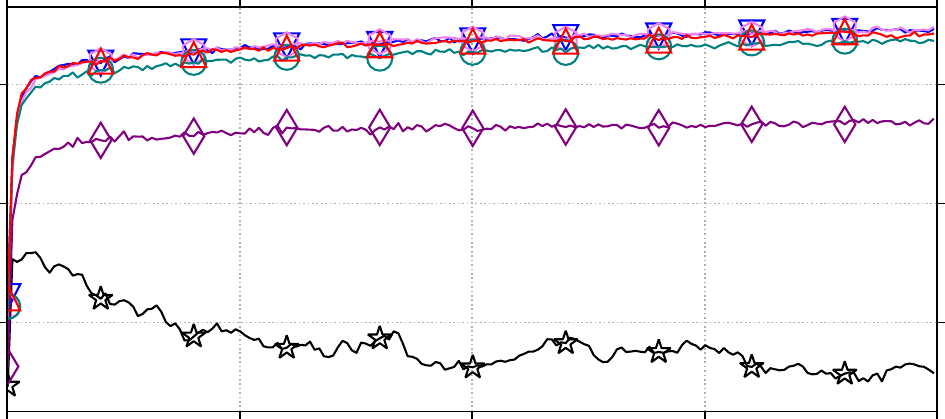}
        \\
        \raisebox{3\normalbaselineskip}[0pt][0pt]{\rotatebox[origin=c]{90}{
\footnotesize Istella - NDCG@10
}}
        &
        \includegraphics[scale=0.29]{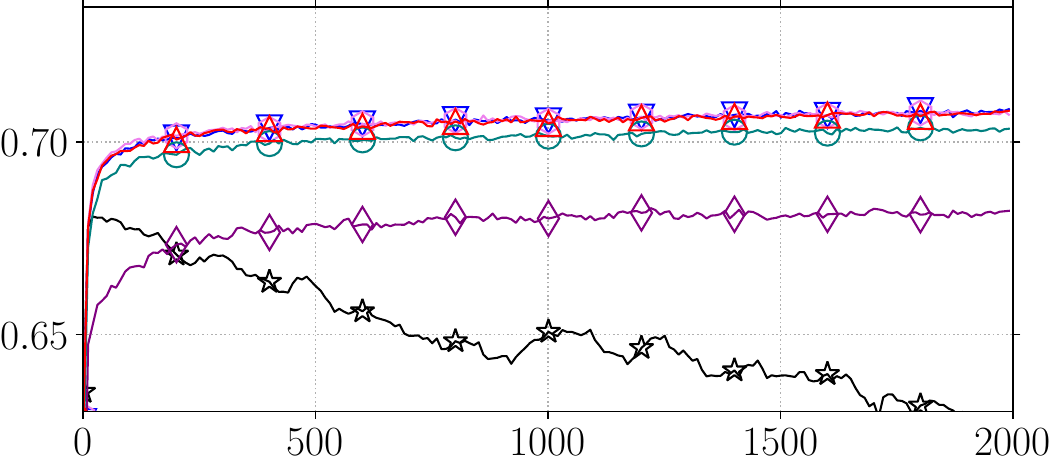}
        &
        \includegraphics[scale=0.29]{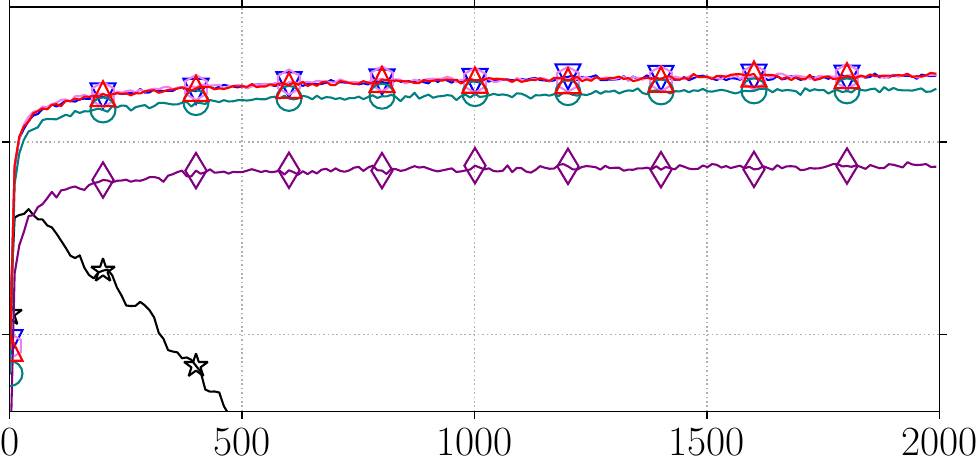}
        &
        \includegraphics[scale=0.29]{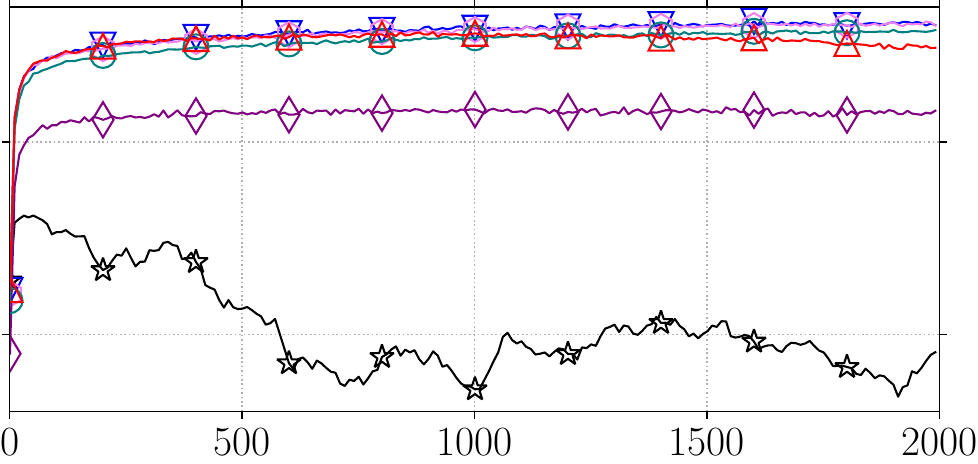}
    \end{tabular}
    \vspace{-1\baselineskip}
    \caption{Performance of different estimators \emph{without} baseline corrections (i.e., $b = 0$) for varying numbers of samples ($N$) per gradient estimate. Y-axis: NDCG@10 on test set; X-axis: training epochs.
    Top-row: MSLR-Web30k; Bottom-row: Istella-S.}
    \label{fig:nobaseline}
    \vspace{-\baselineskip}
\end{figure*}
}

{
\renewcommand{\arraystretch}{0.3}
\setlength\tabcolsep{1pt}
\begin{figure*}[t]
    \centering
    \begin{tabular}{r l l l l}
    & \multicolumn{1}{c}{\footnotesize N = 2}
    & \multicolumn{1}{c}{\footnotesize N = 10}
    & \multicolumn{1}{c}{\footnotesize N = 100}
    & \multicolumn{1}{c}{\footnotesize N = 1000}
    \\
    \raisebox{2.75\normalbaselineskip}[0pt][0pt]{\rotatebox[origin=c]{90}{
\footnotesize MSLR - NDCG@10
}}
        &
        \includegraphics[scale=0.29]{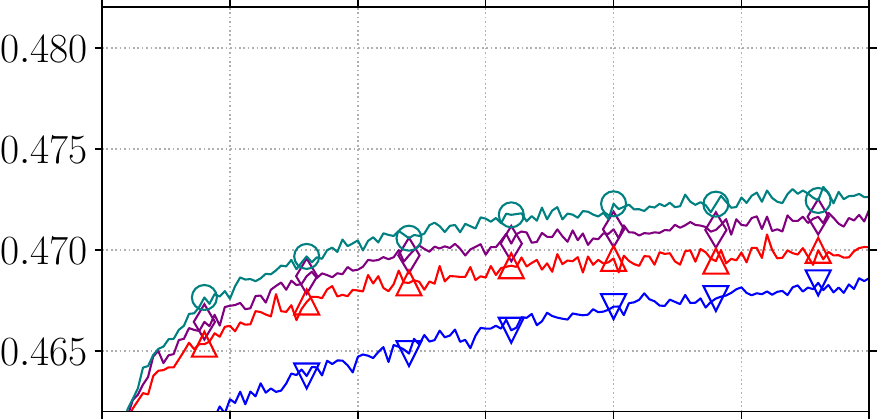}%
        &
        \hspace{0.1mm}%
        \includegraphics[scale=0.29]{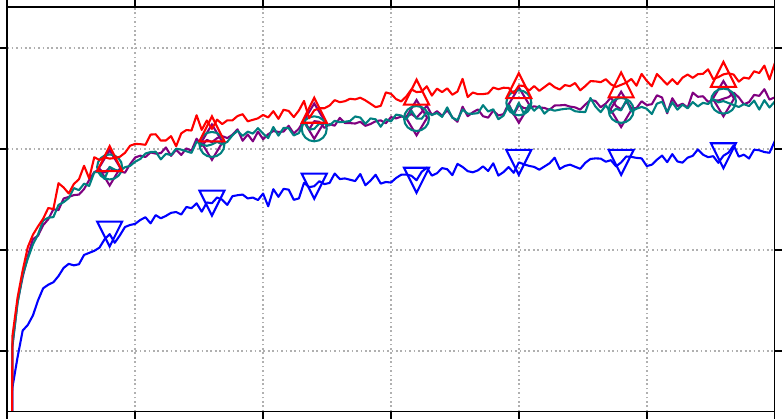}%
        &
        \hspace{0.1mm}%
        \includegraphics[scale=0.29]{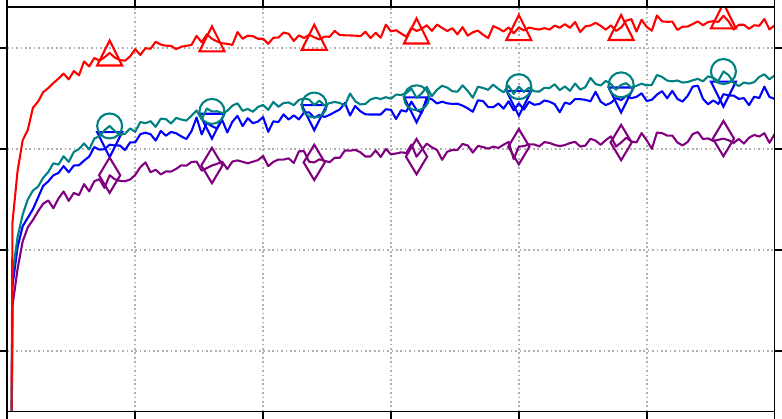}
        &
        \hspace{0.1mm}%
        \includegraphics[scale=0.29]{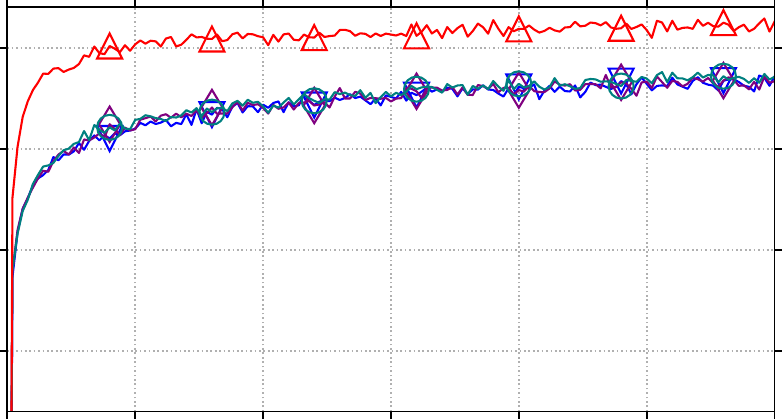}
        \\
        \raisebox{3.1\normalbaselineskip}[0pt][0pt]{\rotatebox[origin=c]{90}{
\footnotesize Istella - NDCG@10
}}
        &
        \hspace{0.93mm}%
        \includegraphics[scale=0.29]{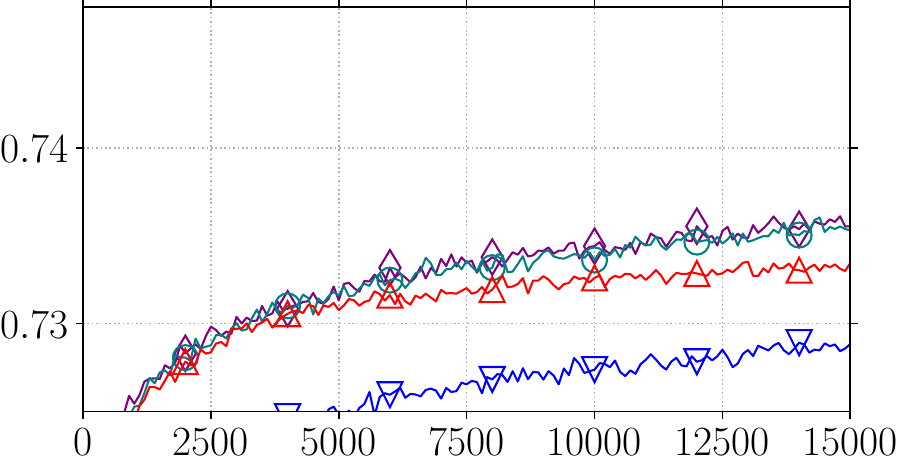}
        &
        \includegraphics[scale=0.29]{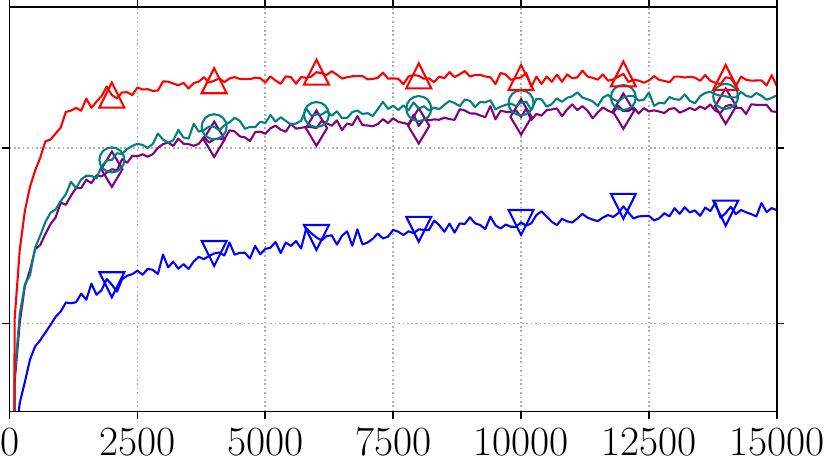}
        &
        \includegraphics[scale=0.29]{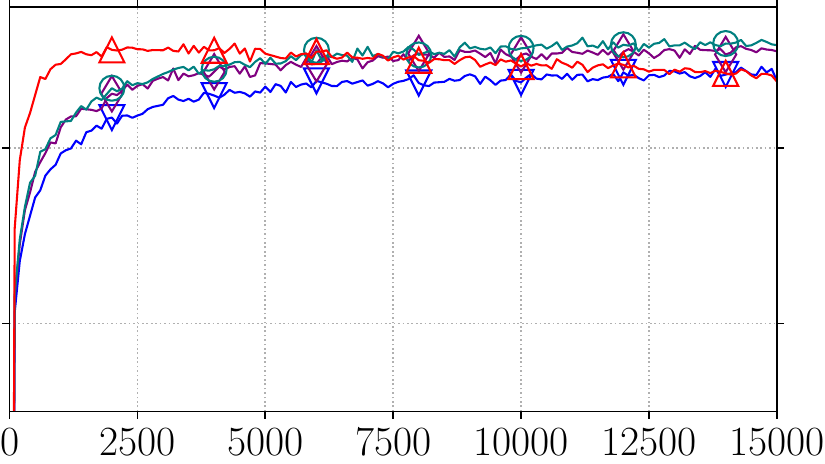}
        &
        \includegraphics[scale=0.29]{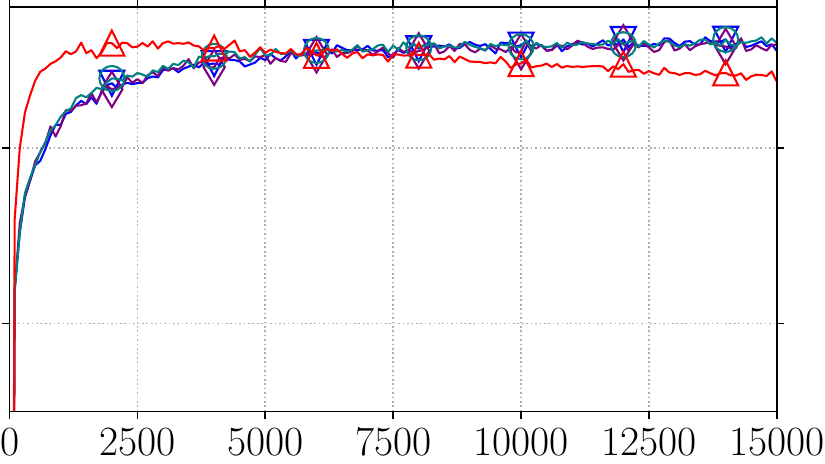}
        \\
        \multicolumn{5}{c}{\includegraphics[scale=0.4]{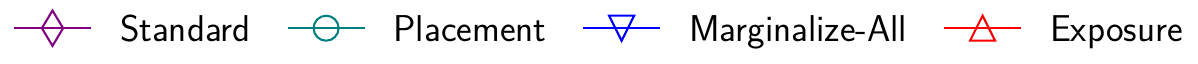}}
    \end{tabular}
   \vspace{-1.2\baselineskip}
    \caption{Performance of different estimators \emph{with} baseline corrections for varying numbers of samples ($N$) per estimate.
    Y-axis: NDCG@10 on test set; X-axis: training epochs.
    Top-row: MSLR-Web30k; Bottom-row: Istella-S.}
    \label{fig:withbaseline}
    \vspace{-\baselineskip}
\end{figure*}
}

\section{Experimental Setup}

Our experiments aim to analyze the effect of our novel \ac{RL} for \ac{LTR} techniques: baseline-corrections, marginalization and exposure-based estimation, in terms of \emph{effectiveness}, i.e., performance at convergence; and \emph{efficiency}, i.e., epochs needed to converge, sample-efficiency and computational costs.
The six methods we compare are:
\begin{enumerate*}[label=(\roman*)]
    \item the \emph{PL-Rank-3} algorithm~\citep{oosterhuis2022} (Eq.~\ref{eq:PLRank}),
    \item the \emph{standard policy-gradient} estimator (Eq.~\ref{eq:estimator:standard}),
    \item the \emph{placement trajectory} estimator (Eq.~\ref{eq:estimator:placement}),
    \item the \emph{marginalize-first} estimator (Eq.~\ref{eq:estimator:margfirst}),
    \item the \emph{marginalize-all} estimator (Eq.~\ref{eq:margallestimator}), and
    \item the \emph{exposure-based} estimator (Eq.~\ref{eq:estimator:exposure}-\ref{eq:estimator:exposureloss}).
\end{enumerate*}
Additionally, we also consider all methods without baseline corrections (PL-Rank never has baseline corrections).

We use two public LTR benchmark datasets: MSLR-Web30k~\citep{qin2013introducing} and Istella-S LETOR~\citep{lucchese2016istella}.
Each datasets contains numerous queries with a preselected set of documents per query represented by feature vectors and expert-judged relevance labels for each (preselected) query-document pair.
MSLR-Web30k has 30,000 queries, on average 125 preselected documents per query and 136 features in their representations;
Istella-S has 33,018 queries on average 103 documents per query and 220 features.
We use the standard division of train, validation and test set folds; MSLR comes in 5 different possible folds, we spread our experimental runs equally over them.

The ranking models optimized are three hidden-layer feed-for\-ward neural networks with layer sizes: $[1024,512, 256]$ with ReLU activations~\citep{hara2015relu}; except for PL-Rank which reached better results with a two hidden-layer network with $32$ units each and sigmoid activations.
We applied the Adamax optimizer~\citep{kingma2014adam} with learning rate of $10^{-3}$, batch normalization with momentum $0.999$, dropout with a rate of $0.5$, and a batch-size of 128 queries.

All methods optimize for top-10 rankings with the DCG weighting scheme: $\theta_k = (\log_2(k + 1))^{-1}$~\citep{jarvelin2002cumulated}.
The metric used for evaluating utility is normalized DCG@10 (NDCG@10); we also optimize the other four exposure-based loss functions from Section~\ref{eq:prelim:exposure}, for their evaluation, we consider their values on the held-out test-set.
For $L_\text{KL-distill}$, we choose $\mathbf{\theta}^*$ as the exposure distribution of the ideal ranking (sorting according to labels); for $L_\text{frac-fair}$, we add $0.01$ to zero relevance labels to avoid divisions by zero.
All results are means over 25 independent runs; to test for significant differences, we use a two-sided student's t-test.
Each experiment was run on a single GPU in a cloud setup, to fairly compare computational costs every GPU used was a NVIDIA® Tesla® V100 for NVLink with 64GiB RAM and 32 bit floats. %
We measure computation time of all optimization procedures (while excluding the computation time of evaluation metrics).
Our implementation was done in JAX~\citep{bradbury2018jax}, making extensive use of the RAX framework~\citep{jagerman2022rax}.

\section{Results}

\subsection{The unstable PL-Rank custom gradient}
\label{sec:results:plrank}

We start by considering the effectiveness of PL-Rank and its stability of convergence in particular.
Figure~\ref{fig:nobaseline} displays the performance of PL-Rank and the other estimators with their baseline corrections \emph{disabled} (PL-Rank has no baseline corrections by default).

Clearly, PL-Rank is unable to converge stably and within the first 250 epochs its performance starts to degrade dramatically.
We have verified that this is not overfitting behavior as the same degradation takes place in performance on the training set.
This observation is surprising considering the conclusion of previous work~\citep{oosterhuis2021plrank, oosterhuis2022}, however, \citet{oosterhuis2021plrank} only considered 40 epochs, and on this scale degradation does not yet occur.
Interestingly, \citet{oosterhuis2022} ran for a set number of minutes, which means the number of epochs varied with the dataset and number of samples $N$, for the few methods that run many epochs (250+) degradation did appear to occur.
Oosterhuis is both an authors of this work and the inventor of PL-Rank and he has extensively verified that our implementation is correct.
We also found that a smaller network with sigmoid activations was able to stabilize the initial learning curve; likely because this leads to much smaller predicted ranking score values.
There are some differences with previous implementations, we ran PL-Rank on GPU and only used 32 bit floats, whereas the original PL-Rank works ran it on CPU and with 64 bit floats~\citep{oosterhuis2021plrank, oosterhuis2022}.\footnote{ \url{https://github.com/HarrieO/2022-SIGIR-plackett-luce}}
The 64 bit precision of the original works was chosen to counter instability issues, thus, it is likely that this difference exacerbated such issues in our experiments.
Nevertheless, we consider our comparison fair as all methods are tested in the same conditions, we note that none of the other estimators has stability issues while also being constrained to 32 bit floats.
Importantly, PL-Rank stores many unnormalized quantities in its intermediate steps, whereas the other methods compute all probabilities in log-space; this can explain why stability issues only occur for PL-Rank.

Therefore, we conclude that \emph{PL-Rank is unable to converge on adequate performance due to its computationally unstable custom gradient}; it appears we are the first to observe the severity of this issue due to only using 32 bit floats and running for a considerably larger number of epochs than previous work.

{
\renewcommand{\arraystretch}{0.92}
\setlength\tabcolsep{1.75pt}
\begin{table*}[t]
\caption{
Performance (NDCG@10 on test-set) reached by different estimators for varying numbers of samples per gradient estimate $N$ on two different datasets.
Model parameters were selected from 15,000 training epochs to maximize NDCG@10 on the validation set.
Results are averages over 25 runs; standard deviation in parentheses.
Statistically significant differences between the exposure-based estimator and the highest baseline are indicated by $^{\blacktriangle}$/$^{\blacktriangledown}$
($p < 0.01$, two-sided student's t-test).
}
\label{tab:main}
\vspace{-\baselineskip}
\resizebox{\linewidth}{!}{
\begin{tabular}{l l l l l l l l l l}
\toprule
\small \# Samples& \multicolumn{1}{c}{ \small $N=2$ }& \multicolumn{1}{c}{ \small $N=5$ }& \multicolumn{1}{c}{ \small $N=10$ }& \multicolumn{1}{c}{ \small $N=25$ }& \multicolumn{1}{c}{ \small $N=50$ }& \multicolumn{1}{c}{ \small $N=100$ }& \multicolumn{1}{c}{ \small $N=250$ }& \multicolumn{1}{c}{ \small $N=500$ }& \multicolumn{1}{c}{ \small $N=1000$ }\\
\midrule
& \multicolumn{9}{c}{\small \it MLSR-Web30k} \\
\midrule
\small PL-Rank & {  0.4544} \tiny (0.0030) & {  0.4631} \tiny (0.0036) & {  0.4496} \tiny (0.0034) & {  0.4474} \tiny (0.0032) & {  0.4446} \tiny (0.0029) & {  0.4557} \tiny (0.0029) & {  0.4538} \tiny (0.0029) & {  0.4513} \tiny (0.0032) & {  0.4493} \tiny (0.0031)  \\
\small Standard & {  0.4728} \tiny (0.0048) & {  0.4769} \tiny (0.0025) & {  0.4778} \tiny (0.0030) & {  0.4772} \tiny (0.0063) & {  0.4789} \tiny (0.0026) & {  0.4773} \tiny (0.0066) & {  0.4788} \tiny (0.0027) & {  0.4783} \tiny (0.0026) & {  0.4785} \tiny (0.0029)  \\
\small Placement & { \bf 0.4735} \tiny (0.0027) & {  0.4753} \tiny (0.0055) & {  0.4780} \tiny (0.0028) & {  0.4785} \tiny (0.0030) & {  0.4787} \tiny (0.0032) & {  0.4788} \tiny (0.0030) & {  0.4788} \tiny (0.0029) & {  0.4785} \tiny (0.0024) & {  0.4789} \tiny (0.0025)  \\
\small Marg.-All & {  0.4695} \tiny (0.0027) & {  0.4734} \tiny (0.0029) & {  0.4756} \tiny (0.0027) & {  0.4771} \tiny (0.0026) & {  0.4773} \tiny (0.0027) & {  0.4778} \tiny (0.0031) & {  0.4786} \tiny (0.0028) & {  0.4785} \tiny (0.0031) & {  0.4785} \tiny (0.0029)  \\
\small Exposure & {  0.4719}$^{\blacktriangledown}$\!\! \tiny (0.0030) & { \bf 0.4773} \tiny (0.0028) & { \bf 0.4792} \tiny (0.0029) & { \bf 0.4812}$^{\blacktriangle}$\!\! \tiny (0.0024) & { \bf 0.4808}$^{\blacktriangle}$\!\! \tiny (0.0026) & { \bf 0.4815}$^{\blacktriangle}$\!\! \tiny (0.0025) & { \bf 0.4813}$^{\blacktriangle}$\!\! \tiny (0.0027) & { \bf 0.4812}$^{\blacktriangle}$\!\! \tiny (0.0026) & { \bf 0.4812}$^{\blacktriangle}$\!\! \tiny (0.0022)  \\
\midrule
& \multicolumn{9}{c}{\small \it Istella-S LETOR} \\
\midrule
\small PL-Rank & {  0.7223} \tiny (0.0048) & {  0.7314} \tiny (0.0016) & {  0.7123} \tiny (0.0036) & {  0.7102} \tiny (0.0024) & {  0.7074} \tiny (0.0015) & {  0.7226} \tiny (0.0031) & {  0.7191} \tiny (0.0029) & {  0.7148} \tiny (0.0078) & {  0.7132} \tiny (0.0064)  \\
\small Standard & { \bf 0.7370} \tiny (0.0013) & {  0.7408} \tiny (0.0012) & {  0.7431} \tiny (0.0010) & {  0.7449} \tiny (0.0010) & { \bf 0.7455} \tiny (0.0010) & { \bf 0.7461} \tiny (0.0011) & {  0.7461} \tiny (0.0009) & {  0.7464} \tiny (0.0010) & {  0.7462} \tiny (0.0009)  \\
\small Placement & {  0.7366} \tiny (0.0012) & {  0.7413} \tiny (0.0010) & {  0.7437} \tiny (0.0011) & {  0.7455} \tiny (0.0011) & { \bf 0.7455} \tiny (0.0011) & { \bf 0.7461} \tiny (0.0012) & { \bf 0.7462} \tiny (0.0009) & { \bf 0.7466} \tiny (0.0008) & {  0.7463} \tiny (0.0009)  \\
\small Marg.-All & {  0.7301} \tiny (0.0014) & {  0.7345} \tiny (0.0012) & {  0.7376} \tiny (0.0012) & {  0.7417} \tiny (0.0011) & {  0.7434} \tiny (0.0010) & {  0.7450} \tiny (0.0013) & {  0.7459} \tiny (0.0011) & {  0.7465} \tiny (0.0010) & { \bf 0.7466} \tiny (0.0010)  \\
\small Exposure & {  0.7343}$^{\blacktriangledown}$\!\! \tiny (0.0040) & { \bf 0.7428}$^{\blacktriangle}$\!\! \tiny (0.0011) & { \bf 0.7450}$^{\blacktriangle}$\!\! \tiny (0.0008) & { \bf 0.7457} \tiny (0.0012) & { \bf 0.7455} \tiny (0.0011) & {  0.7458} \tiny (0.0010) & {  0.7457} \tiny (0.0013) & {  0.7462} \tiny (0.0010) & {  0.7462} \tiny (0.0008)  \\
\bottomrule
\end{tabular}

}
\vspace{-\baselineskip}
\end{table*}
}

\subsection{Marginalization and baseline corrections}

To analyze the effect of our novel marginalization and baseline corrections, we consider Figure~\ref{fig:nobaseline} and Figure~\ref{fig:withbaseline} which shows the learning curves of the estimators with baseline corrections disabled and enabled, respectively.
Additionally, Table~\ref{tab:main} displays (test-set) performance reached when selecting model parameters that reached the best validation set performance (baseline corrections enabled).

In Figure~\ref{fig:nobaseline}, the effect of marginalization can be observed in the absence of baseline corrections.
Ignoring PL-Rank, the differences between methods matches our expectation on both datasets and for all values of $N$: the standard estimator performs worst, the placement does much better, and then marginalization reaches even higher performance.
There is barely a difference between the marginalize-first, marginalize-all and exposure-based estimators; this is expected for the latter two as they are equivalent without baseline corrections.
The similarity with the marginalize-first estimator is more surprising, and indicates that the first placement is the most important for gradient estimation; due to its high similarity with the marginalize-all estimator and to avoid cluttering our results, we omit it from further comparisons.
From our observations in Figure~\ref{fig:nobaseline}, we conclude that \emph{in the absence of baseline corrections, marginalization substantially improves performance}.

In contrast, Figure~\ref{fig:withbaseline} shows us that the differences between methods becomes more complicated with baseline corrections.
Ubiquitously, we see an enormous performance gain from baseline-corrections for all methods (except PL-Rank), to the point that the Y-axis of Figure~\ref{fig:withbaseline} barely overlaps with that of Figure~\ref{fig:nobaseline}.
However, there appears to be an interaction between marginalization and baseline corrections as not every method receives the same gain.
In particular, marginalize-all now has the poorest performance despite being the best when baseline corrections are disabled; the standard estimator has high performance that is very similar to the placement estimator, despite being consistently the worst without baseline corrections (cf.\ Figure~\ref{fig:nobaseline}).
We attribute this difference due to the standard baseline correction better matching its estimator, i.e., the gradient is based on entire rankings while the correction is the expected reward over entire rankings; in contrast, the marginalize-all gradient is based on partial-rankings conditioned on the prefixes of a sampled ranking, while its correction is unconditioned.
When the number of samples $N=1000$, any meaningful differences between all estimators except PL-Rank and exposure-based have disappeared.
From these observations, we draw the conclusion that \emph{baseline corrections greatly improve performance but can have unexpected interactions with marginalization}, as a result, marginalization may decrease performance if baseline corrections are also applied.

\subsection{Exposure-based RL for LTR}
From Figure~\ref{fig:withbaseline} and Table~\ref{tab:main}, we see that with $N \geq 5$, the performance of the exposure-based estimator outperforms all other methods in learning speed, i.e., epochs needed to reach high performance, and performance at convergence in several settings, i.e., on MSLR it reaches substantially higher NDCG@10 than any other method with $N\geq 100$.
However, it is outperformed by the standard and placement estimators when $N = 2$; the baseline correction of the exposure-based estimator is its exposure distribution, and thus, it appears that a sufficient number of samples is required for this correction to be accurate enough.
Mathematically, its baseline correction is the only difference with the marginalize-all estimator, thus the large difference in performance between them reveals that the exposure-based correction is much more effective.

Furthermore, on Istella, the NDCG@10 of the exposure-based method decreases after 2500 epochs and $N \geq 100$, we verified that this is due to overfitting by observing continued steady improvements in performance on the training-set.\footnote{We note that this overfitting can be avoided with early stopping, as was done for the results in Table~\ref{tab:other}. We consider the fact that overfitting is reached in substantially fewer epochs than other methods to be strong evidence that it is considerably faster, which we deem to be a very desirable trait for a gradient-estimation method.}
Finally, Table~\ref{tab:main} also performed a significance test on the differences with the exposure-based method, we find that when $N\geq5$ it always has competitive performance (marginally close to the highest) and in many cases reaches significant improvements.
In particular, we only find significant decreases for $N=2$; and the exposure-based estimator converges at significantly better performance on MSLR with $N\geq25$ and on Istella with $ 10 \geq N \geq 5$.
Figure~\ref{fig:withbaseline} shows us that even when no better optimum is found, the exposure-based estimator reaches competitive performance substantially faster than other estimators, requiring around 33\% of epochs (${\sim}2500$ instead of ${\sim}7500$).

We conclude that the \emph{exposure-based estimator provides significant improvements in performance at convergence and learning speed} over other \ac{RL} for \ac{LTR} estimators, as long as $N\geq5$.
Overall, our results show that the choice of baseline correction is the most important for effectiveness, and in our comparison, exposure-based is evidently the best baseline correction choice.

\begin{figure}[t]
\centering
    \includegraphics[scale=0.287]{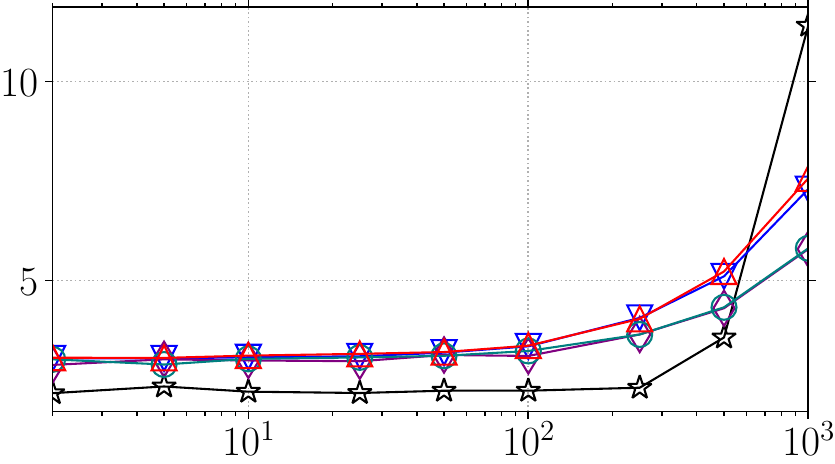}
    \includegraphics[scale=0.287]{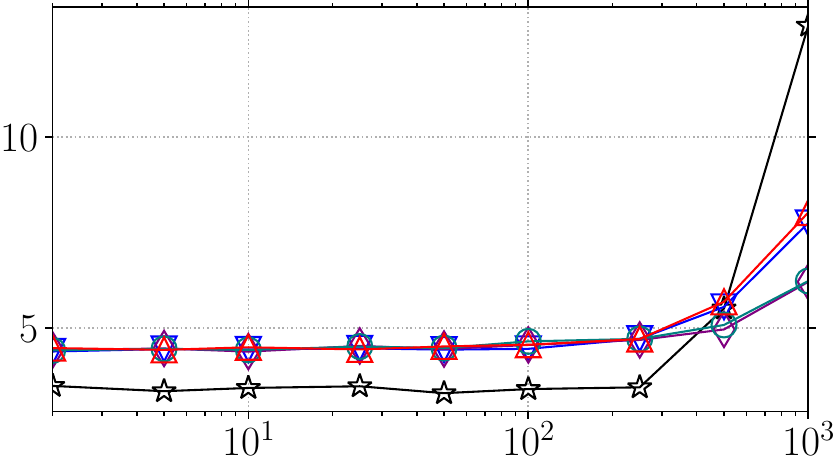}
    \includegraphics[scale=0.33]{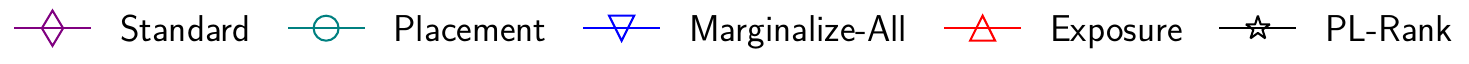}
    \vspace{-\baselineskip}
    \caption{Seconds (Y-axis) used per training epoch by different estimators for varying numbers of samples $N$ (X-axis).
    Left: MSLR-Web30k; Right: Istella-S LETOR.}
    \label{fig:timing}
    \vspace{-1.3\baselineskip}
\end{figure}

{
\renewcommand{\arraystretch}{0.005}
\setlength\tabcolsep{0pt}
\begin{figure*}[t]
    \centering
    \begin{tabular}{r l l l l}
    & \multicolumn{1}{c}{\hspace{1mm}\footnotesize $L_\text{frac-fair}$ (Eq.~\ref{eq:loss:fracfair})}
    & \multicolumn{1}{c}{\hspace{3mm}\footnotesize $L_\text{prod-fair}$ (Eq.~\ref{eq:loss:prodfair})}
    & \multicolumn{1}{c}{\hspace{3mm}\footnotesize $L_\text{KL-fair}$ (Eq.~\ref{eq:loss:KLfair})}
    & \multicolumn{1}{c}{\hspace{2mm}\footnotesize $L_\text{KL-distill}$ (Eq.~\ref{eq:loss:KLdistill})}
    \\
    \raisebox{2.5\normalbaselineskip}[0pt][0pt]{\rotatebox[origin=c]{90}{
\footnotesize MSLR
}}\;
        &
        \includegraphics[scale=0.29]{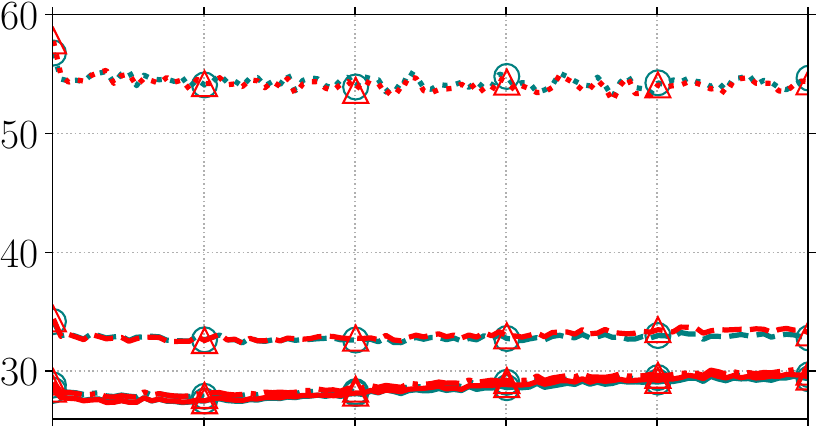}
        &
        \includegraphics[scale=0.29]{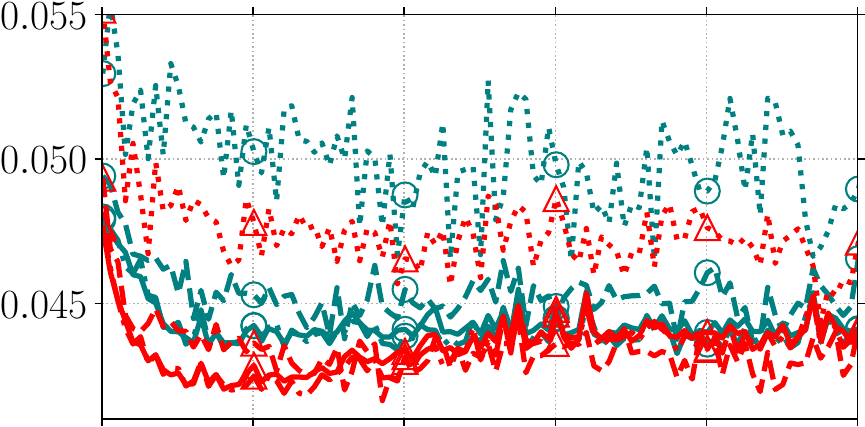}
        &
        \includegraphics[scale=0.29]{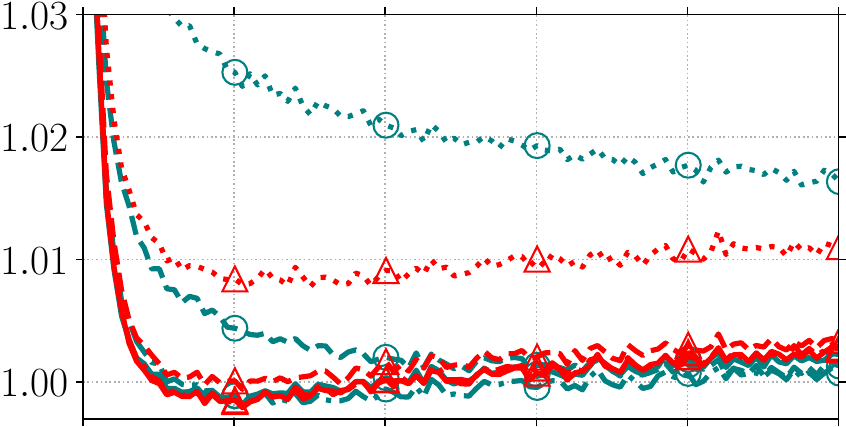}
        &
        \includegraphics[scale=0.29]{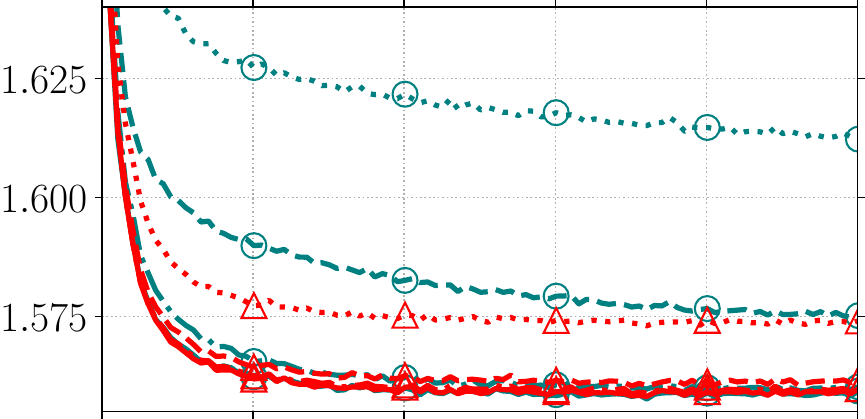}
        \\
        \raisebox{3.2\normalbaselineskip}[0pt][0pt]{\rotatebox[origin=c]{90}{
\footnotesize Istella
}}\;
        &
        \includegraphics[scale=0.29]{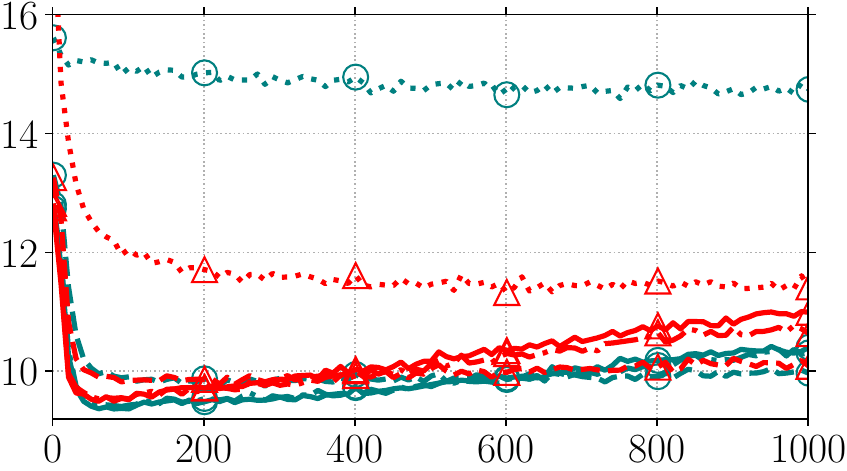}
        &
        \hspace{0.93mm}%
        \includegraphics[scale=0.29]{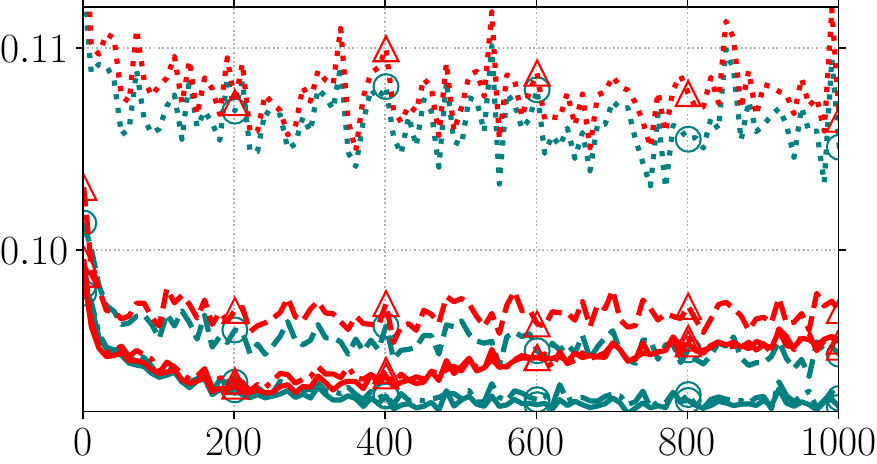}
        &
        \hspace{0.93mm}%
        \includegraphics[scale=0.29]{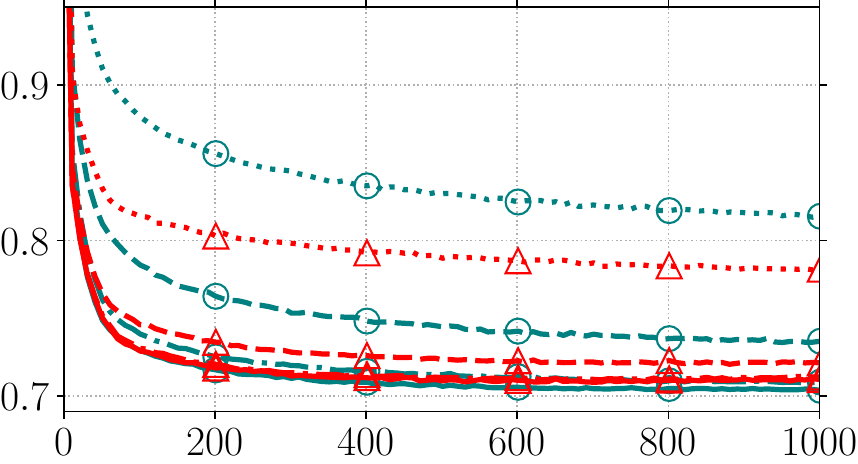}
        &
        \hspace{0.93mm}%
        \includegraphics[scale=0.29]{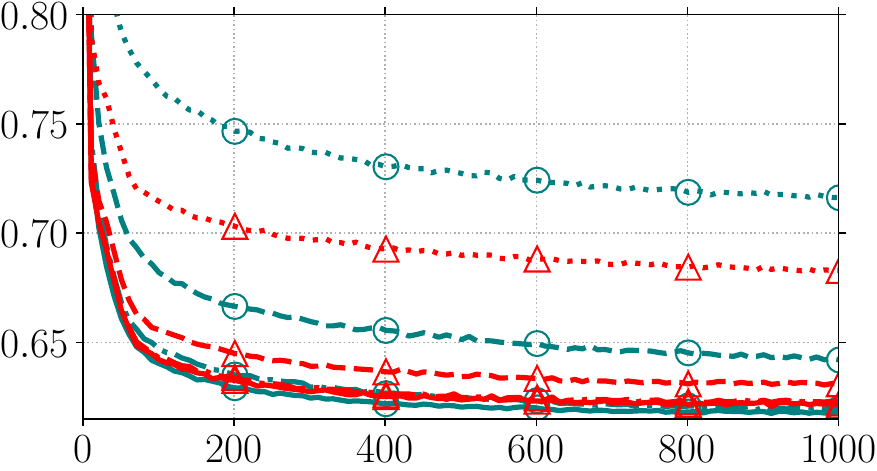}
        \\
        \multicolumn{5}{c}{\includegraphics[scale=0.4]{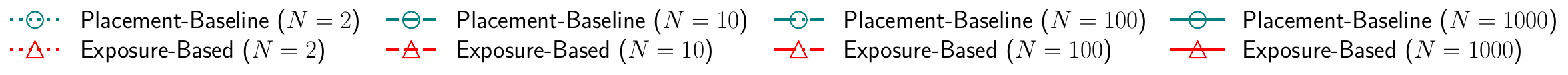}}
    \end{tabular}
    \vspace{-1.2\baselineskip}
    \caption{Comparison of exposure-based estimator and baseline estimator for optimizing various non-traditional exposure-based loss functions. Y-axis: Mean loss value on test set, X-axis: Training epochs.
    Top-row: MSLR-Web30k; Bottom-row Istella-S.}
    \label{fig:otherlosses}
    \vspace{-\baselineskip}
\end{figure*}
}

\subsection{Computational costs of exposure-based LTR}

Figure~\ref{fig:timing} displays the mean number of seconds used for a single training epoch for different estimators and values of $N$.
Interestingly, when $N\leq100$, time per epoch is stable for all estimators; a result of how GPU computation executes matrix operations.
When $N > 100$, times increase and differences between methods shift.

PL-Rank stands out for having the lowest time per epoch when $N\leq100$, this is a result of its clever custom gradient which was designed to minimize computation time~\citep{oosterhuis2022}.
In light of our conclusions in Section~\ref{sec:results:plrank}, it becomes clear that PL-Rank has made an extreme trade-off between numerical stability and computation speed.
Unfortunately, its trade-off is too extreme to be worthwhile; as effectiveness is completely sacrificed for computational efficiency.
Surprisingly, when $N > 500$, PL-Rank is the least computationally-efficient method, we attribute this to the scatter additions that the algorithm requires, which do not scale well on GPU.

Unexpectedly, despite requiring more computations, marginalization does not noticeably increase time per epoch when $N\leq 100$.
Even at $N = 1000$, its increase appears marginal as it adds less than 2.5 seconds per epoch.
In Table~\ref{tab:main}, we see marginal improvements in NDCG@10 when increasing $N$ beyond $100$, additionally, Figure~\ref{fig:withbaseline} shows that the learning curves of the exposure-based estimator are virtually indistinguishable when $N=100$ and $N=1000$.
Thus, for the exposure-based estimator, there appears to be no reason to choose an $N > 100$ for effectiveness, and when $N\leq 100$ it is at least as efficient as the other estimators.
Therefore, we conclude that \emph{the significant improvements in effectiveness of the exposure-based estimator do not come with a decrease in computational efficiency} when using GPUs for gradient computation.

{
\setlength\tabcolsep{1.5pt}
\begin{table}[t]
\caption{
Test-set loss values achieved by exposure-based and placement (baseline) approaches for varying numbers of samples.
Results are averages over 25 independent runs (1,000 epochs each); standard deviation in parentheses.
}
\label{tab:other}
 \vspace{-\baselineskip}
\resizebox{\linewidth}{!}{
\begin{tabular}{l l l l l l}
\toprule
\multicolumn{2}{l}{ \small \# Samples } & \multicolumn{1}{c}{ \small $N=2$ } & \multicolumn{1}{c}{ \small $N=10$ } & \multicolumn{1}{c}{ \small $N=100$ } & \multicolumn{1}{c}{ \small $N=1000$ }\\
\midrule
& & \multicolumn{4}{c}{\small \it MLSR-Web30k} \\
\midrule
\multirow{2}{*}{\small \it $L_\text{frac-fair}$} & \small Placement & {  54.73} \tiny (03.51) & {  32.72} \tiny (02.06) & {  28.23} \tiny (01.95) & { \bf 27.43} \tiny (01.93)  \\
& \small Exposure & { \bf 54.48} \tiny (03.05) & { \bf 32.57} \tiny (01.89) & { \bf 28.06} \tiny (02.19) & {  27.61} \tiny (01.90)  \\
\cmidrule{3-6}
\multirow{2}{*}{\small \it $L_\text{prod-fair}$} & \small Placement & {  0.048} \tiny (0.014) & {  0.046} \tiny (0.015) & { \bf 0.043} \tiny (0.013) & {  0.045} \tiny (0.014)  \\
& \small Exposure & { \bf 0.047} \tiny (0.010) & { \bf 0.044} \tiny (0.014) & { \bf 0.043} \tiny (0.013) & { \bf 0.044} \tiny (0.013)  \\
\cmidrule{3-6}
\multirow{2}{*}{\small \it $L_\text{KL-fair}$} & \small Placement & {  1.011} \tiny (0.006) & {  1.000} \tiny (0.006) & { \bf 0.998} \tiny (0.005) & {  1.000} \tiny (0.004)  \\
& \small Exposure & { \bf 1.008} \tiny (0.005) & { \bf 0.999} \tiny (0.005) & { \bf 0.998} \tiny (0.005) & { \bf 0.998} \tiny (0.005)  \\
\cmidrule{3-6}
\multirow{2}{*}{\small \it $L_\text{KL-distill}$} & \small Placement & {  1.598} \tiny (0.005) & {  1.565} \tiny (0.004) & { \bf 1.559} \tiny (0.005) & { \bf 1.558} \tiny (0.005)  \\
& \small Exposure & { \bf 1.573} \tiny (0.005) & { \bf 1.561} \tiny (0.005) & { \bf 1.559} \tiny (0.005) & {  1.559} \tiny (0.005)  \\
\midrule
& & \multicolumn{4}{c}{\small \it Istella-S LETOR} \\
\midrule
\multirow{2}{*}{\small \it $L_\text{frac-fair}$} & \small Placement & {  14.23} \tiny (00.24) & { \bf 09.84} \tiny (00.21) & { \bf 09.48} \tiny (00.18) & { \bf 09.40} \tiny (00.11)  \\
& \small Exposure & { \bf 11.55} \tiny (00.42) & {  09.98} \tiny (00.25) & {  09.53} \tiny (00.14) & {  09.50} \tiny (00.13)  \\
\cmidrule{3-6}
\multirow{2}{*}{\small \it $L_\text{prod-fair}$} & \small Placement & { \bf 0.105} \tiny (0.006) & { \bf 0.095} \tiny (0.003) & { \bf 0.093} \tiny (0.001) & { \bf 0.092} \tiny (0.001)  \\
& \small Exposure & {  0.106} \tiny (0.005) & { \bf 0.095} \tiny (0.002) & { \bf 0.093} \tiny (0.001) & {  0.093} \tiny (0.001)  \\
\cmidrule{3-6}
\multirow{2}{*}{\small \it $L_\text{KL-fair}$} & \small Placement & {  0.794} \tiny (0.002) & {  0.723} \tiny (0.002) & { \bf 0.706} \tiny (0.002) & { \bf 0.703} \tiny (0.002)  \\
& \small Exposure & { \bf 0.774} \tiny (0.003) & { \bf 0.719} \tiny (0.002) & {  0.709} \tiny (0.002) & {  0.708} \tiny (0.002)  \\
\cmidrule{3-6}
\multirow{2}{*}{\small \it $L_\text{KL-distill}$} & \small Placement & {  0.698} \tiny (0.002) & {  0.629} \tiny (0.001) & { \bf 0.616} \tiny (0.001) & { \bf 0.615} \tiny (0.002)  \\
& \small Exposure & { \bf 0.674} \tiny (0.002) & { \bf 0.628} \tiny (0.002) & {  0.621} \tiny (0.002) & {  0.619} \tiny (0.002)  \\
\bottomrule
\end{tabular}

}
\vspace{-\baselineskip}
\end{table}
}

\subsection{Non-traditional exposure-based losses}
Finally, we consider the performance of the exposure-based estimator for non-traditional loss functions, specifically, the four exposure-based losses from Section~\ref{eq:prelim:exposure} (excluding utility $U$).
Originally, we intended to compare with the existing approach suggested by \citet{oosterhuis2021plrank}, where partial gradients are computed of the loss w.r.t.\ the exposure of a document, and the gradient values are then used as relevance labels for the PL-Rank algorithm.
However, since we found PL-Rank highly unstable (see Section~\ref{sec:results:plrank}), we replace PL-Rank with the placement estimator for our baseline approach.

Figure~\ref{fig:otherlosses} displays the learning curves for the non-traditional losses; and Table~\ref{tab:other} shows the test-set loss values reached by selecting the best model weights based on the validation-set loss.
In Figure~\ref{fig:otherlosses}, we see several cases where the exposure-based estimator learns faster when $10 \geq N \geq 2$, but also other cases where the differences in speed are marginal.
Furthermore, the exposure-based method appears more prone to overfitting for $L_\text{frac-fair}$ and $L_\text{prod-fair}$ losses on Istella.
However, when we consider the loss values reached in Table~\ref{tab:main}, the differences between the methods appear mostly marginal when $N \geq 100$.
Thus, we conclude that \emph{the exposure-based and placement approaches are both effective at optimizing non-traditional loss functions} at comparable levels of effectiveness.

Additionally, we also measured the time per epoch for both methods, where we expected the baseline needs more time due to requiring two gradient computations per update.
However, no meaningful differences in time were measured.
It appears that JAX's just-in-time compilation~\citep{bradbury2018jax} streamlined the baseline so successfully that the extra gradient steps do not have a measurable impact.

In summary, we observe no meaningful differences between the exposure-based and baseline approach in terms of performance and computational costs.
Therefore, we conclude that \emph{the increase in ease-of-implementation provided by the exposure-based estimator does not come at the cost of effectiveness nor efficiency}.

\section{Conclusion}

In this work, we have contributed several new \ac{RL} techniques for \ac{LTR}, while avoiding using custom gradients and focusing on ease-of-implementation, variance reduction and GPU computation.
We proposed \emph{baseline corrections} and a novel \emph{partial marginalization} approach to increase the sample-efficiency of traditional \ac{RL} for \ac{LTR} for relevance maximization.
Moreover, we introduced a novel \emph{exposure-based} estimator which abstracts gradient estimation behind a document-exposure distribution.
Thereby, it integrates well with auto-differentiation software and also considerably simplifies and streamlines the implementation of exposure-based losses.

Our experimental results show that the existing PL-Rank algorithm is extremely unstable when computed with 32 bit precision and unable to reliable converge on adequate performance.
In contrast, baseline corrections and marginalization improve both learning speed and performance at convergence.
However, there are interactions between the choice of baseline correction and marginalization due to which their improvements may not stack straightforwardly.
Our results clearly show that the exposure-based estimator provides the highest effectiveness and efficiency in our comparison, leading to significantly higher NDCG@10 at convergence on the MSLR-Web30k dataset and requiring considerably fewer training epochs on both the MSLR-Web30k and Istella-S datasets.
Additionally, when using GPU computation, the exposure-based estimator has the same computational costs as standard \ac{RL} estimators.
Lastly, it is also effective and efficient at optimizing non-traditional exposure-based \ac{LTR} losses, e.g., for fairness and distillation.

Thereby, we have considerably improved the effectiveness, efficiency and ease-of-implementation of \ac{RL} for \ac{LTR}.
We hope our contributions will stimulate the adoption of \ac{RL} for \ac{LTR} by practitioners.
To support this further, we have added the exposure-based estimator to the publicly available RAX framework~\citep{jagerman2022rax}: \url{https://github.com/google/rax}.
In addition, a standalone implementation is publicly available at: \url{https://github.com/HarrieO/2026-ictir-exposure-LTR}.

\begin{acks}
This research was supported by the Google Visiting Researcher program.
Any opinions, findings and recommendations expressed in this work are those of the authors and are not necessarily shared or endorsed by their respective employers or sponsors.

We thank Don Metzler and Michael Bendersky and our reviewers for useful discussions and constructive feedback.
We thank Oscar Ramirez Milian for his help with testing and perfecting the publicly available implementation of the exposure-based estimator.
\end{acks}

\appendix

\begin{listing}[t]
\begin{minted}[fontsize=\footnotesize]{python}
import jax.numpy as jnp
from jax.lax import stop_gradient
from utils import sample_rankings, cumlogsumexp
# input: [D] vector of scores, [K] vector of exposure, RNG key and N
# output: [D] vector of (estimated) expected exposure per document
def exposure(scores, rank_exposure, rng_key, n_samples):
  K = rank_exposure.shape[0]
  # creates [N, D] matrix of sampled rankings from PL ranking model
  rankings = sample_rankings(rng_key, n_samples, scores)
  ranked_scores = scores[rankings] 
  # [N, K] log denominator of placement prob. per position per sample
  log_denom = cumlogsumexp(ranked_scores, axis=1, reverse=True)[:,:K]
  # [N, K-1] log prob. of sampled ranking up to pos. k (not including)
  log_prefix = ranked_scores[:,:K-1] - log_denom[:,:K-1]
  log_prefix = jnp.cumsum(jnp.pad(log_prefix, ((0,0),(1,0))), axis=1)
  # [N, K, D] placement prob. of every document per position & sample
  prob_in = jnp.exp(scores[None,None,:] - log_denom[:,:,None])
  mask = jnp.zeros_like(prob_in, dtype=bool)
  mask = mask.at[jnp.arange(n_samples)[:,None], jnp.arange(1,K)[None,:],
                 rankings[:,:K-1]].set(True)
  mask = jnp.cumsum(mask, axis=1).astype(bool)
  prob_in = jnp.where(mask, 0, prob_in)
  prob_out = jnp.where(mask[:,-1,:], 0, 1-prob_in[:,-1,:])
  # [N, K, D] placement prob. per position and outside rankings
  prob_all = jnp.concat((prob_in, prob_out[:,None,:]), axis=1)
  log_prefix_all = jnp.concat((log_prefix, log_prefix[:,-1:]), axis=1)
  rank_exposure_all = jnp.pad(rank_exposure, (0,1))
  # [N, K, D] loss value used for computing policy gradient
  loss_K_D = prob_all + stop_gradient(prob_all)*log_prefix_all[:,:,None]
  exposure_prob_prod = prob_all * rank_exposure_all[None,:,None]
  # [N, D] exposure of document per sample
  sample_exposure = jnp.sum(exposure_prob_prod, axis=1)
  # [D] expected exposure of document estimated from samples
  mean_exposure = jnp.mean(sample_exposure, axis=0)
  # [N, D] leave-one-out baseline corrections
  baseline = mean_exposure[None,:] - sample_exposure/n_samples
  baseline *= n_samples / (n_samples-1)
  # [N, K, D] baseline-corrected rewards for every doc., pos., sample
  rewards = stop_gradient(rank_exposure_all[None,:,None]-baseline[:,None,:])
  loss = jnp.mean(jnp.sum(rewards * loss_K_D, axis=1), axis=0)
  # trick: value is expected exposure but gradient comes from loss
  return stop_gradient(mean_exposure) + loss - stop_gradient(loss)
\end{minted}
\caption{Implementation of the exposure-based estimator.}
\label{listing:estimator-corrected}
\end{listing}

\section{Addendum: Correction of Exposure Gradient}

During the preparation of the camera-ready version of this work, we realized an error was made in the derivation of the gradient of the exposure-based estimator, due to which Eq.~\ref{eq:estimator:exposure} is incorrect.
This mistake stems from a difference in how baselines should be added to the exposure estimate (Eq.~\ref{eq:estimator:exposurevalue}).
In contrast with the utility estimates in Section~\ref{sec:method:relevance} where one can substract the baseline per position placement, for exposure estimates this is incorrect:
\begin{equation}
\begin{split}
    \theta_{d|\pi,q} &= \mathbb{E}_{y \sim \pi(q)}\big[ \theta_{\text{rank}(d \mid y)} \big]
    = b_d + \mathbb{E}_{y \sim \pi(q)}\big[ \theta_{\text{rank}(d \mid y)} - b_d \big]
    \\ &\not=
    b_d + \sum_{k=1}^K (\theta_k - b_d) P(y_k = d \mid \pi).
\end{split}
\end{equation}
This does not work because $\sum_{k=1}^K P(y_k = d \mid \pi) \leq 1$ and therefore the baseline corrections inside the expectation do not cancel the outside baseline value.
The key insight here is that there are $K+1$ placement possibilities for a document: it can be placed on one of the $K$ positions or not be placed in the ranking at all, therefore:
$P(d \not\in y \mid \pi, q) + \sum_{k=1}^K P(y_k = d \mid \pi) = 1$.
Accordingly, we have:
\begin{equation}
    \theta_{d|\pi,q}
    =
    b_d - b_dP(d \not\in y \mid \pi, q) + 
    \sum_{k=1}^K (\theta_k - b_d) P(y_k = d \mid \pi, q).
\end{equation}
The probability of not placing a document $d$ in a ranking can be estimated using our marginalization strategy:
\begin{align}
& P(d \not\in y \mid \pi, q) = \!\!\!\! \sum_{y_{1:K-1}}\!\!\! \pi\mleft(y_{1:K-1} \! \mid q \mright)\mleft(1 - \pi\mleft(d \mid y_{1:K-1} \mright)\mright)\mathds{1}\mleft[d \not\in y_{1:K-1} \mright]
\nonumber \\[-1.8ex] &\quad\; \approx
\frac{1}{N} \sum_{i=1}^N \pi\mleft(y_{1:K-1}^{(i)} \mid q \mright)\mleft(1 - \pi\mleft(d \mid y_{1:K-1}^{(i)}  \mright)\mright)\mathds{1}\mleft[d \not\in y_{1:K-1}^{(i)} \mright],
\end{align}
and thus the gradient of the exposure-based estimator by:
\begin{align}
\frac{\delta }{\delta f}
    \theta_{d|\pi,q}
    &\approx 
    \frac{1}{N} \sum_{i=1}^N \bigg( \sum_{k=1}^K \mleft(\theta_k - b^{(i)}_d \mright)\bigg(\mleft[ \frac{\delta }{\delta f} \pi\mleft(d \mid y_{1:k-1}^{(i)} \mright)\mright]
    \nonumber \\[-1ex]
    & \qquad\qquad + 
    \pi\mleft(d \mid y_{1:k-1}^{(i)} \mright)\mleft[ \frac{\delta }{\delta f} \log \pi\mleft(y_{1:k-1}^{(i)} \mid q \mright) \mright]\bigg)
    \nonumber \\[-0.6ex]
    &\quad - b^{(i)}_d \mathds{1}\mleft[d \not\in y_{1:K-1}^{(i)} \mright] \bigg(\mleft[ \frac{\delta }{\delta f} \mleft( 1 - \pi\mleft(d \mid y_{1:K-1}^{(i)} \mright) \mright) \mright]
    \\[-0.6ex]
    & \quad\;\; + 
    \mleft(1 - \pi\mleft(d \mid y_{1:K-1}^{(i)}\mright) \mright)\mleft[ \frac{\delta }{\delta f} \log \pi\mleft(y_{1:K-1}^{(i)} \mid q \mright) \mright]\bigg)
    \bigg),
     \nonumber \\[-1ex]
    b^{(i)}_d &=
\frac{1}{N-1}\sum_{j=1}^N \mathds{1}\mleft[i \not= j\mright] \sum_{k = 1}^K \pi\mleft(d \mid y_{1:k-1}^{(j)}\mright) \theta_k.
\nonumber
\end{align}
Listing~\ref{listing:estimator-corrected} presents a possible implementation.
Unfortunately, the experiments could not be repeated with the corrected estimator before the camera-ready deadline.
We are working on follow-up study that includes it in its experimental comparison.

\bibliographystyle{ACM-Reference-Format}
\bibliography{references}

\end{document}